\documentclass[twoside]{article}

\usepackage[accepted]{aistats2024}
\usepackage[round]{natbib}
\PassOptionsToPackage{numbers}{natbib}
\usepackage[utf8]{inputenc} 
\usepackage[T1]{fontenc}    

\usepackage{url}            
\usepackage{booktabs}       
\usepackage{amsfonts}       
\usepackage{nicefrac}       
\usepackage{microtype}      
\usepackage{xcolor}         
\usepackage{algorithm}
\usepackage{algorithmic}
\usepackage{subcaption}
\usepackage{setspace}
\usepackage{enumitem}
\usepackage{comment}
\usepackage{booktabs}
\usepackage{multirow}

\usepackage{amsmath,bm}
\usepackage{amssymb}
\usepackage{mathtools}
\usepackage{amsthm}
\usepackage{xspace}
\usepackage{upgreek}

\usepackage[framemethod=tikz]{mdframed}

\definecolor{ocre}{RGB}{243,102,25}
\definecolor{mygray}{RGB}{243,243,244}

\usepackage{hyperref}       
\hypersetup{%
    pdfborder = {0 0 0},
    colorlinks,
    citecolor=blue,
    linkcolor=blue,
}

\usepackage[capitalize,noabbrev]{cleveref}
\usepackage{empheq}

\theoremstyle{plain}

\theoremstyle{definition}

\theoremstyle{remark}

\newcommand{\E}{\operatornamewithlimits{\mathbb{E}}} 
\newcommand{\bmu}{\bm{\mu}}

\newcommand{\bx}{\bm{x}}

\newcommand{\bW}{\bm{W}}
\newcommand{\bsigma}{\bm{\sigma}}
\newcommand{\n}{\mathsf{n}}
\newcommand{\repsilon}{\upepsilon}

\newcommand{\rxi}{\upxi}
\newcommand{\rtheta}{\uptheta}
\newcommand{\z}{\textsf{z}}
\newcommand{\w}{w}
\newcommand{\W}{W}
\newcommand{\m}{\mathsf{m}}
\newcommand{\rM}{\mathsf{B}}
\newcommand{\rbx}{\bm{\mathsf{x}}}

\newcommand{\inc}{\mathtt{inc}}
\newcommand{\cv}{\mathtt{cv}}
\newcommand{\dual}{\mathtt{joint}}
\newcommand{\naive}{\mathtt{naive}}
\newcommand{\ens}{\mathtt{ens}}

\newcommand{\vi}{}

\newcommand{\diag}{\mathrm{diag}}

\newcommand{\lr}{}
\newcommand{\V}{\operatornamewithlimits{\mathbb{V}}}

\newcommand{\vp}{q_{w}(z)}

\newcommand{\tw}{{\mathcal{T}_{w}(\epsilon)}}

\newcommand{\bsf}[1]{\mathsf{\bm{#1}}}

\begin{document}

%

%

\twocolumn[

\aistatstitle{Joint control variate for faster black-box variational inference
}

\aistatsauthor{ Xi Wang \And Tomas Geffner \And Justin Domke}

\aistatsaddress{Manning College of Information and Computer Sciences,\\University of Massachusetts Amherst \\ \texttt{\{xwang3,tgeffner,domke\}@cs.umass.edu}} ]

\begin{abstract}
Black-box variational inference performance is sometimes hindered by the use of gradient estimators with high variance. This variance comes from two sources of randomness: Data subsampling and Monte Carlo sampling. While existing control variates only address Monte Carlo noise, and incremental gradient methods typically only address data subsampling, we propose a new "joint" control variate that \emph{jointly} reduces variance from both sources of noise. This significantly reduces gradient variance, leading to faster optimization in several applications.
\end{abstract}

\section{INTRODUCTION}

Black-box variational inference (BBVI)~\citep{hoffman2013stochastic, titsias2014doubly, ranganath2014black, kucukelbir2017automatic, blei2017variational} is a popular alternative to Markov Chain Monte Carlo (MCMC) methods. The idea is to posit a variational family and optimize it to be close to the posterior, using only "black-box" evaluations of the target model (either the density or gradient).
This is typically done by minimizing the KL-divergence using stochastic optimization methods with unbiased gradient estimates.
Often, this allows the use of data subsampling, which greatly speeds-up optimization with large datasets.

The BBVI optimization problem is often called "doubly-stochastic"~\citep{titsias2014doubly,salimbeni2017doubly}, as the gradient estimation has two sources of randomness: Monte Carlo sampling from the variational distribution, and data subsampling from the full dataset.
Because of the doubly-stochastic nature, one common challenge for BBVI is the variance of the gradient estimates: If this is high, it forces small stepsizes, leading to slow optimization convergence~\citep{nemirovski2009robust,bottou2018optimization}. 

Numerous methods exist to reduce the "Monte Carlo" noise that comes from drawing samples from the variational distribution~\citep{miller2017reducing,roeder2017sticking,geffner_using,geffner2020approximation,boustati2020amortized}. These can typically be seen as creating an approximation of the objective for which the Monte Carlo noise can be integrated exactly. This approximation can then be used to define a control variate—a zero mean random variable that is negatively correlated with the original gradient estimator. These methods can sometimes be used with data subsampling, essentially by creating different approximations for each datum. However, they are only able to reduce \emph{per-datum} Monte Carlo noise—they do not reduce subsampling noise itself. This is critical, as subsampling noise is often the dominant source of gradient variance (Sec.~\ref{sec:var_where}).

For (non-BBVI) optimization problems with \emph{only} subsampling noise, there are numerous incremental gradient methods, that "recycle" previous gradient evaluations to speed up convergence \citep{roux2012stochastic, shalev2013stochastic, johnson2013accelerating, defazio2014saga, defazio2014finito}. However, with few exceptions (Sec.~\ref{sec:related}) these methods do not address Monte Carlo noise and, due to how they rely on efficiently maintaining running averages, cannot be applied to doubly-stochastic problems.

This paper presents a method that \emph{jointly} controls Monte Carlo and subsampling noise. The idea is to create approximations of the target for each datum, where the Monte Carlo noise can be integrated exactly. The method maintains running averages of the \emph{approximate} gradients, with noise integrated, overcoming the challenge of applying incremental gradient ideas to doubly-stochastic problems. The method addresses both forms of noise and also \emph{interactions} between them.
Experiments with variational inference on a range of probabilistic models show that the method yields lower variance gradients and significantly faster convergence than existing approaches.

\vspace{-2pt}
\section{BACKGROUND: BLACK-BOX VARIATIONAL INFERENCE}
\vspace{-2pt}

Given a probabilistic model
$p(x, z) = p(z) \prod_{n=1}^{N}p(x_n \mid z)$
and observed data $x_1,\ldots,x_N$, variational inference's goal is to find a tractable distribution $\vp$
to approximate the (often intractable) posterior $p(z\mid x)$ over the latent variable $z \in \mathbb{R}^d$. BBVI achieves this by finding the parameters $w$ that minimize the KL-divergence from $\vp$ to $p(z \mid x)$, equivalent to minimizing the negative Evidence Lower Bound (ELBO)
\begin{equation}
\label{eq:elbo}
f^{\vi}(w) = -\E_\n\E_{q_w(\z)}\bigg[N \log p(x_\n \mid \z) + \log p(\z)\bigg]
- \mathbb{H}(w),
\end{equation}
where $\mathbb{H}(w)$ denotes the entropy of $q_{w}$.


The expectation with respect to $\z$ in \cref{eq:elbo} is typically intractable. Thus, BBVI methods rely on stochastic optimization with unbiased gradient estimates, usually based on the score function method~\citep{williams1992simple} or the reparameterization trick~\citep{kingma2014auto,rezende2014stochastic, titsias2014doubly}.
The latter is often the method of choice due to the fact that it often yields estimators with lower variance~\citep{kucukelbir2017automatic,xu2019variance}.
The idea is to define a fixed base distribution $s(\epsilon)$ and a deterministic transformation $\tw$ such that for $\repsilon \sim s$, we have $\tw \sim q_w$. Then, the objective in \cref{eq:elbo} can be re-written as
\begin{equation}
    f(w)^{\vi} = \E_\n\E_{\repsilon} f(w; \n, \epsilon),
\end{equation}
where  
\begin{multline}
\label{eqn:reparam_elbo}
    f(w; n, \epsilon) = -N \log p(x_n \mid \mathcal{T}_w(\epsilon) ) - \log p(\mathcal{T}_w(\epsilon))\\- \mathbb{H}(w).
\end{multline}
The "naive" gradient estimate is obtained by drawing a random $n$ and $\epsilon$, and evaluating
\begin{equation}\label{eq:g_naive_n_eps}
    g_{\naive}(w; n, \epsilon) = \nabla f(w; n, \epsilon).
\end{equation}
Since this only requires point-wise evaluations of $\log p$ and its gradient, it can be applied to a diverse range of models, including those with complex and non-conjugate likelihoods. And by subsampling data, it can be used with large datasets, which may be challenging for traditional methods like MCMC~\citep{hoffman2013stochastic, kucukelbir2017automatic}. However, the effectiveness of this strategy depends on the gradient estimator's variance; if it is too large, then very small step sizes will be required, slowing convergence.

\section{GRADIENT VARIANCE IN BBVI} 
\label{sec:var_where}

Let $\V_{\n, \repsilon}[\nabla f(w;\n,\repsilon)]$ denote the variance of the naive estimator from Eq.~\eqref{eq:g_naive_n_eps}.\footnote{When $\z$ is a vector, we use $\V[\z]=\mathrm{tr}\  \mathbb{C}[\z]$.} The two sources of variance correspond to data subsampling ($n$) and Monte Carlo noise ($\epsilon$). It is natural to ask how much variance each of these sources contributes.

\begin{figure}[t]
    \centering
        \includegraphics[width=0.48\textwidth]{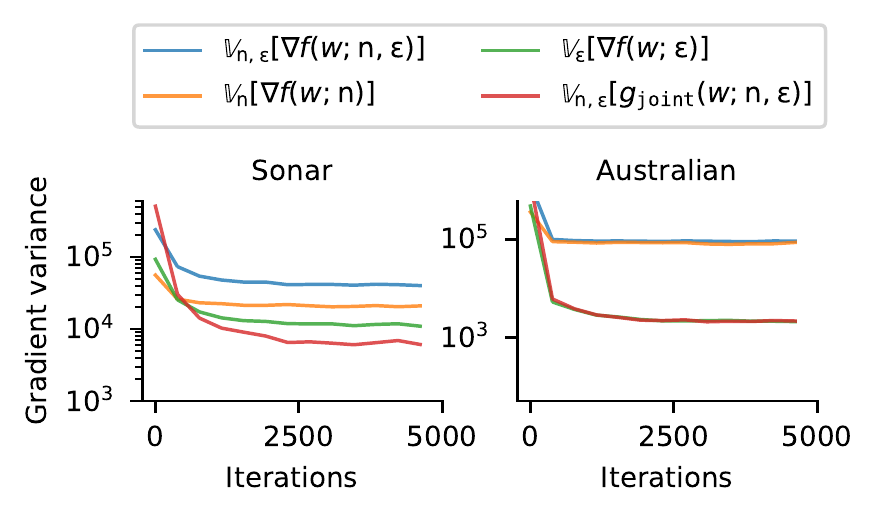}
    \caption{
    \textbf{The contributions of subsampling and Monte Carlo noise vary by problem. The proposed joint estimator reduces both.}
    Orange lines denote variance from data subsampling ($n$), and green lines denote Monte Carlo noise variance ($\epsilon$). We use a batch size of 5. For the Sonar dataset, both sources show similar scales. For the Australian dataset, subsampling noise dominates. Regardless, our proposed gradient estimator $g_\dual$ (red line, Eq.~\eqref{eq:g_dual}) mitigates subsampling noise and controls MC noise, aligning closely with or below green lines (i.e. the variance without data subsampling) in both datasets.
    } 
    \label{fig:small_var_decompo}
\end{figure}

Let $f(w;n)=\E_\repsilon f(w;n,\repsilon)$ be the objective for a single datum $n$ with Monte Carlo noise integrated out.
Similarly, let $f(w;\epsilon)=\E_\n f(w;\n,\epsilon)$ be the objective for a fixed $\epsilon$ evaluated on the full dataset.
In Fig.~\ref{fig:small_var_decompo} and Table.~\ref{table:model_variances}, we do a single run of BBVI using our proposed gradient estimator (described below).
Then, for each iteration on that single optimization trace, we estimate the variance of $\nabla f(w;\n,\repsilon)$, $\nabla f(w;\repsilon)$, and $\nabla f(w;\n)$.
We do this for multiple tasks, described in detail in Sec.~\ref{sec:experiment}.
For later reference, we also include the joint estimator developed below.

The amount of variance contributed by each source is task-dependent. But in many tasks considered, subsampling noise is larger than Monte Carlo noise. This is problematic since computing $f(w;\repsilon)$ requires looping over the full dataset, eliminating any benefit of subsampling. These results also illustrate the limitations of any approach that only handles a single source of noise: No control variate applied to each datum can do better than $\nabla f(w;n)$, while no incremental-gradient-type method can do better than $\nabla f(w;\epsilon)$. 



\begin{table}[t]
\scalebox{0.85}{
\footnotesize
\centering
\begin{tabular}{llll}
\toprule
Task    &  $\V_{\n,\repsilon}[\nabla f(w;\n,\repsilon)]$ & $\V_{\n}[\nabla f(w;\n)]$ & $\V_\repsilon[\nabla f(w;\repsilon)]$ \\
\midrule
Sonar     & $4.04\times 10^{4}$  & $2.02\times 10^{4}$  & $1.16\times 10^{4}$  \\
Australian      & $9.16\times 10^{4}$  & $8.61\times 10^{4}$  & $2.07\times 10^{3}$  \\
MNIST     & $4.21\times 10^{8}$  & $3.21\times 10^{8}$  & $1.75\times 10^{4}$  \\
PPCA     & $1.69\times 10^{10}$ & $1.68\times 10^{10}$ & $3.73\times 10^{7}$  \\
Tennis   & $9.96\times 10^{7}$  & $9.59\times 10^{7}$  & $8.56\times 10^{4}$  \\
MovieLens & $1.78 \times 10^{9}$ & $1.69 \times 10^{9}$  & $1.75 \times 10^{6}$ \\
\bottomrule
\end{tabular}}
\caption{BBVI gradient variance decomposition across tasks, computed at the optimization endpoint. With a batch size of 5, step size of $5\times 10^{-4}$ for Sonar and Australian, a batch size of 100, step size of $1\times 10^{-2}$ for others. We generally observe subsampling noise $\V_\n[\nabla f(w;\n)]$ dominates MC noise $\V_\repsilon[\nabla f(w;\repsilon)]$.}
\label{table:model_variances}
\end{table}

\section{VARIANCE REDUCTION FOR STOCHASTIC OPTIMIZATION}\label{sec:background_cv}

This section introduces existing methods of variance reduction for stochastic optimization problems with a single source of gradient variance and their applicability to doubly-stochastic settings.

\subsection{Monte Carlo sampling and approximation-based control variates}\label{sec:PSO}

Suppose we sum over the full dataset in each iteration. Then the objective from Eq.~\ref{eqn:reparam_elbo} becomes $f(w) = \E_\repsilon f(w;\repsilon)$. A gradient can easily be estimated by sampling $\epsilon$. 
Previous work ~\citep{Paisley2012VariationalBI,tucker2017rebar,grathwohl2018backpropagation,boustati2020amortized} has proposed to reduce the variance using a \emph{control variate}~\citep{robert1999monte}: A (zero-mean) random variable $c(w; \repsilon)$ negatively correlated with the gradient estimator and defining the new estimator
\begin{equation}\label{eq:pso_grad}
    g(w;\epsilon)=\nabla f(w;\epsilon) + c(w;\epsilon).
\end{equation}
The hope is that $c(w;\epsilon) \approx \nabla f(w) - \nabla f(w;\epsilon)$ approximates the noise of the original estimator, which can lead to large reductions in variance and thus more efficient and reliable inference.

A general way to construct control variates involves using an approximation function $\tilde f\approx f$ for which the expectation $\E_\repsilon \tilde{f}(w, \repsilon)$ is available in closed-form~\citep{miller2017reducing, geffner2020approximation}. Then, the control variate is defined as $c(w;\epsilon) =\E_\rxi \nabla \tilde{f}(w;\rxi) - \nabla \tilde{f}(w; \epsilon)$, and the estimator from Eq.~\eqref{eq:pso_grad} becomes
\begin{equation}\label{eq:pso_cv_grad}
g(w; \epsilon)=\nabla f(w;\epsilon) + \E_\rxi \nabla \tilde{f}(w; \rxi) - \nabla \tilde{f}(w; \epsilon).
\end{equation}
The better  $\tilde f$ approximates $f$, the lower the variance of this estimator tends to be. (For a perfect approximation, the variance is zero.) A popular choice for $\tilde f$ is a quadratic function as the expectation of a quadratic under a Gaussian is tractable. The quadratic can be learned~\citep{geffner2020approximation} or obtained through a second-order Taylor expansion~\citep{miller2017reducing}. 

In doubly-stochastic problems of the form $f(w;n,\epsilon)$, data $n$ is subsampled as well as $\epsilon$. While the above control variate has typically been used without subsampling, it can be adapted to the doubly-stochastic setting by developing an approximation $\tilde f(w;n,\epsilon)$ to $f(w;n,\epsilon)$ for each datum $n$. This leads to the control variate $\E_\rxi \nabla \tilde{f}(w;n, \rxi) - \nabla \tilde{f}(w; n, \epsilon)$ and gradient estimator
\vspace{-5pt}
\begin{multline}\label{eq:g_cv_w_n_eps}
    g_{\cv}(w; n, \epsilon)= \nabla f(w;n,\epsilon) \\ +  \underbrace{\E_\rxi \nabla \tilde{f}(w;n, \rxi) - \nabla \tilde{f}(w; n, \epsilon)}_{\text{zero mean control variate } c_{\cv}(w;n,\epsilon)}.
\end{multline}
Note, however, that such a control variate cannot reduce subsampling noise. Even if $\tilde f(w;n,\epsilon)$ were a \emph{perfect} approximation
there would still be gradient variance due to $n$ being sampled randomly. Using the law of total variance, one can show that
\begin{multline}\label{eq:g_cv_var_final}
\V[g_{\cv}] = \E_{\n}\V_{\repsilon}[\nabla f(w;n,\epsilon) -\nabla \tilde{f}(w; \n, \repsilon)] \\
+ \V_\n[\nabla f(w;\n)]. 
\end{multline}
(See Appendix.~\ref{sec:cv_grad} for a proof.) While the first term on the right-hand side can be made arbitrarily small if $\tilde f$ is close to $f$, the second term is irreducible. Fig.~\ref{fig:sonar_aus_grad_var} and Table~\ref{table:model_variances} show that this subsampling variance is typically substantial, and may be orders of magnitude larger than Monte Carlo variance. When this is true, this type of control variate can only have a limited effect on overall gradient variance.

\vspace{-5pt}
\subsection{Data subsampling and incremental gradient methods}
\label{sec:IGO}
\vspace{-5pt}

Now consider an objective $f(w) = \E_\n f(w;\n)$, where  $\n$ is uniformly distributed on $\{1,\ldots,N\}$ and there is no Monte Carlo noise. While one can compute the exact gradient by looping over $n$, this is expensive when $N$ is large. A popular alternative is to use stochastic optimization by drawing a random $\n$ and using the estimator $\nabla f(w;\n)$. Alternatively, \emph{incremental gradient} methods~\citep{roux2012stochastic,shalev2013stochastic,johnson2013accelerating,defazio2014finito,gower2020variance} can lead to faster convergence. While details vary by algorithm, the basic idea of these methods is to "recycle" previous gradient evaluations to reduce randomness. SAGA \citep{defazio2014saga}, for instance, stores $w^n$ for the most recent iteration where $f(w;n)$ was evaluated and takes a step
\begin{equation}
\label{eq:SAGAupdate}
w \leftarrow w - \lambda \left( \nabla f(w;n) + \E_\m \nabla f(w^{\m}; \m) - \nabla f (w^n; n) \right),
\end{equation}
where $\lambda$ is the step size. The expectation over $m$ is tracked efficiently using a running average, so the cost per iteration is independent of $N$. This update rule can be interpreted as regular stochastic gradient descent using the naive estimator $\nabla f(w;n)$ along with a control variate, i.g.
\begin{equation}
g(w;n) = \nabla f(w;n) + \underbrace{\E_\m \nabla f(w^\m; \m) - \nabla f(w^n; n)}_{\text{zero mean control variate}}.
\label{eq:saga_as_cv}
\end{equation}
When $w^m \approx w$, the first and last terms in Eq.~\eqref{eq:saga_as_cv} will approximately cancel, leading to a gradient estimator with much lower variance.

We now consider a doubly-stochastic objective $f(w;n,\epsilon)$. In principle, one might compute the estimator from Eq.~\eqref{eq:saga_as_cv} for each value of $\epsilon$, i.e. use the gradient estimator 
\begin{multline}\label{eq:g_inc_w_n_eps}
    g_{\inc}(w;n, \epsilon) = \nabla f_n(w;n, \epsilon) \\ +  \underbrace{\E_{\m} \nabla f(w^{\m};\m, \epsilon) - \nabla f (w^{n};n, \epsilon)}_{\text{zero mean control variate}~  c_\inc(w;n,\epsilon)}.
\end{multline}
One issue with this is that it does not address Monte Carlo noise. It can be shown that the variance is
\begin{multline}\label{eq:g_inc_var_final}
\V[g_{\inc}]= \E_{\repsilon}\V_{\n}[\nabla f(w;\n,\repsilon) -\nabla f (w^{\n};\n, \repsilon)] \\
+ \V_{\repsilon}[\nabla f(w;\repsilon)].
\end{multline}
(See Appendix~\ref{sec:inc_grad} for a proof.) Since the second term above is irreducible, the variance does not go to zero even when all the stored parameters $w^n$ are to the current parameters. Intuitively, this estimator cannot do better than evaluating the objective on the full dataset for a random $\epsilon$.

But there is an even larger issue: $g_\inc$ \emph{cannot be implemented efficiently}. The value of $\nabla f(w^\n;n, \epsilon)$ depends on $\epsilon$, which is resampled at each iteration. Therefore, it is not possible to efficiently maintain $\E_{\m} \nabla f(w^{\m};\m, \epsilon)$ (as needed by Eq.~\eqref{eq:g_inc_w_n_eps}) as a running average. The only general strategy is to compute this by looping over the full dataset in each iteration, eliminating the computational benefit of subsampling. For some models with special structures (e.g.\ log-linear models), it is possible to efficiently maintain the needed running average~\citep{wang2013variance,zheng2018lightweight}, but this can only be done in special cases with model-specific derivations, breaking the universality of BBVI.


\subsection{Ensembles of control variate}

It can be valuable to ensemble multiple control variates. For example, \citep{geffner_using} combined control variates that reduced Monte Carlo noise \citep{miller2017reducing} with one that reduced subsampling noise \citep{wang2013variance} (for a special case where $g_\inc$ is tractable). While this approach can be better than either control variate alone, it does not reduce \emph{joint} variance. To see this, consider a gradient estimator that uses a convex combination of the two above control variates. For any $\beta \in (0,1)$ write
\begin{multline}\label{eq:c_combo}
g_\ens(w;n,\epsilon) = \nabla f(w; n, \epsilon)  \\ + \underbrace{\beta  c_{\cv}(w;n, \epsilon) + (1 - \beta) c_{\inc}(w;n, \epsilon)}_{\text{zero mean control variate}~ c_{\ens}(w;n,\epsilon)}.
\end{multline}
Even if both $c_\cv$ and $c_\inc$ are "perfect" (i.e. $\tilde f(w;n,\epsilon)=f(w;n,\epsilon)$ and $w^n=w$ for all $n$), then the variance is
\begin{equation}
    \V[ g_\ens ] = \beta^2 \V_{\n}[\nabla{f}(w;\n)] + (1-\beta)^2 \V_{\repsilon}[\nabla f(w;\repsilon)].
\end{equation}
(See Appendix~\ref{sec:combo_grad} for a proof.) So, even in this idealized scenario, such an estimator cannot reduce variance to zero. Lastly, as $g_\ens$ relies on $c_\inc$, it also faces the computational efficiency issue of $g_\inc$, making it impractical in general problems.

\begin{figure}[t]
    \centering
    \includegraphics[width=0.48\textwidth]{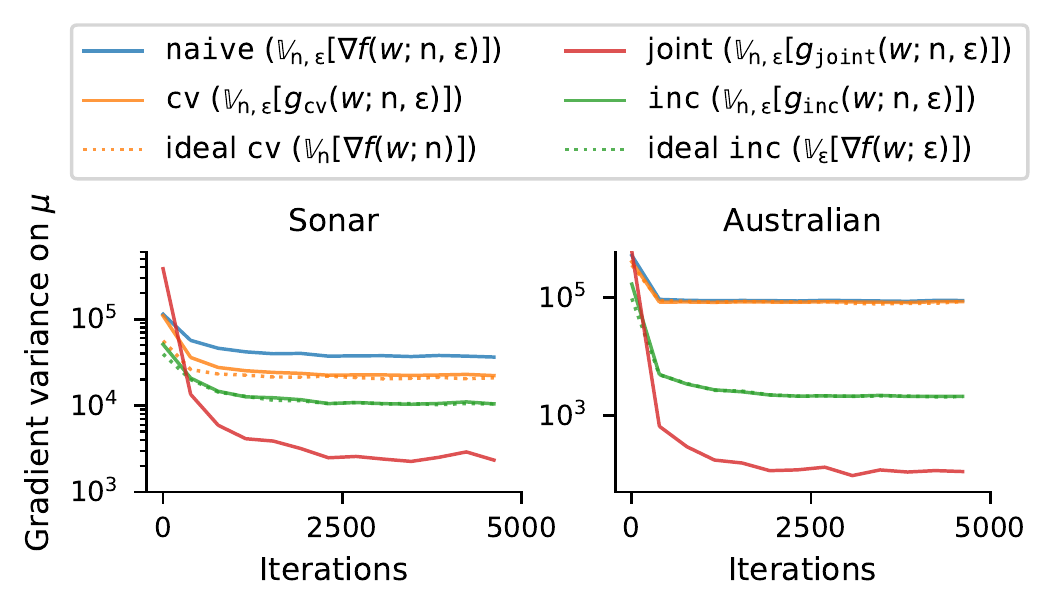}
    \caption{\textbf{In practice, \texttt{cv} and \texttt{inc} reduce variance nearly as much as theoretically possible. The \texttt{joint} estimator variance is lower than these bounds.} The $\naive$ gradient estimator (Eq.~\eqref{eq:g_naive_n_eps}) is the baseline, while the $\cv$ estimator (Eq.~\eqref{eq:g_cv_w_n_eps}) controls the Monte Carlo noise, the $\inc$ estimator (Eq.~\eqref{eq:g_inc_w_n_eps}) controls for subsampling noise, and the proposed $\dual$ estimator (Eq.~\eqref{eq:g_dual}) controls for both. The variance of $\cv$ and $\inc$, as is shown in Eq.~\eqref{eq:g_cv_var_final} and Eq.~\eqref{eq:g_inc_var_final} are lower-bounded by the dotted lines, while $\dual$ is capable of reducing the variance to significantly lower values, leading to better and faster convergence (first two grids in Fig.~\ref{fig:big_results}).}
    \label{fig:sonar_aus_grad_var}
    \vspace{-10pt}
\end{figure}

\section{JOINT CONTROL VARIATE}

We now introduce the \textit{joint control variate}, a new approach for controlling the variance of gradient estimators for BBVI, which \emph{jointly} reduces both subsampling noise from $\n$ and Monte Carlo noise from $\epsilon$. In order to construct such control variate, we take two steps:
\begin{enumerate}[leftmargin=*,topsep=0pt,parsep=0pt]
\item Create an approximation $\tilde f(w;n,\epsilon)$ of the true objective $f(w;n,\epsilon)$, designed so that the expectation $\E_\repsilon \nabla \tilde{f}(w;n,\repsilon)$ can easily be computed for any datum $n$ \citep{miller2017reducing, geffner2020approximation}. A common strategy for this is a Taylor-expansion—replacing $f$ with a low-order polynomial. If the base distribution $s(\epsilon)$ is simple, the expectation $\E_\repsilon[\nabla \tilde f(w; n, \repsilon)]$ may be available in closed-form.
\item Inspired by SAGA~\citep{defazio2014saga}, maintain a table $\W=\{\w^1,\ldots,\w^N \}$ with $w^n \in \mathbb{R}^D$ that stores the variational parameters at the last iteration each of the data points $x_1, \cdots, x_N$ were accessed, along with a running average of gradient estimates evaluated at the stored parameters, denoted by $G$. Unlike SAGA, however, this running average is for the gradients of the \emph{approximation} $\tilde f$, with the Monte Carlo noise $\epsilon$ integrated out, i.e. $G = \E_\n \E_\repsilon \nabla \tilde f(w^\n; \n, \repsilon)$. In practice, we initialize $W$ using a single epoch of optimization with the $\naive$ estimator.
\end{enumerate}
Intuitively, as optimization nears the solution, the parameters $w$ tend to change slowly, meaning the entries $w^n$ in $W$ will tend to become close to the current iterate $w$. So if $\tilde f$ is a good approximation of the true objective, we may expect $\nabla f(w;n,\epsilon)$ to be close to $\nabla \tilde f(w^n;n,\epsilon)$, meaning the two will be strongly correlated. However, thanks to the running average $G$, the full expectation of $\nabla \tilde f(w^n;n,\epsilon)$ is available in closed-form. This leads to our proposed gradient estimator
\begin{multline}
\label{eq:g_dual}
    g_{\dual}(w; n, \epsilon) = \nabla f(w; n, \epsilon)  \\  + \underbrace{\E_\m \E_\rxi \nabla \tilde{f}(w^\m;\m,\rxi) - \nabla \tilde{f}(w^n; n, \epsilon)}_{\text{zero mean control variate }}.
\end{multline}


The running average $G = \E_\n \E_\repsilon \nabla \tilde f(w^\n; \n, \repsilon)$ can be cheaply maintained through optimization, since a single value $w^n$ changes per iteration and $\E_\repsilon \nabla \tilde f(w;n,\repsilon)$ is known in closed form.
The variance of the proposed gradient estimator is
\begin{equation}
    \V[g_\dual] =\V_{\repsilon,\n}[\nabla f(w; \n, \repsilon)  - \nabla \tilde{f}(w^\n; \n, \repsilon)] \label{eq:dualvarline2}.
\end{equation}
This shows that the variance of $g_\dual$ can be arbitrarily small, only limited by how close $\tilde{f}$ is to $f$ and how close the stored values $w^n$ are to the current parameters $w$. This is in contrast with the variance achieved by typical control variates or incremental gradient methods, which are unable to reduce both sources of variance jointly. In fact, as shown in Eq.~\eqref{eq:g_cv_var_final} and Eq.~\eqref{eq:g_inc_var_final}, these methods, even in ideal scenarios, are provably unable to produce estimators with zero variance, as they can only handle a single source of gradient noise.


Alg.~\ref{alg:final} illustrates how the joint gradient estimator can be used for black-box variational inference. The same idea could also be applied more generally to doubly-stochastic objectives in other domains. A generic version of the algorithm and an example of how it can be applied for generalized linear models with Gaussian dropout on the feature is shown in Appendix.~\ref{sec:generic_opt}. 

\paragraph{Memory overhead}
Like SAGA, our method requires $\Theta(ND)$ storage for the parameter table $W$. However, it is easy to create analogous methods based on other incremental gradient methods. In Appendix.~\ref{appendix:SVRG}, we develop an analogous method based on SVRG~\citep{johnson2013accelerating} which only requires $\Theta(D)$ storage. Our empirical evaluation shows that its performance is comparable to the SAGA version. However, it has an extra hyperparameter that controls the frequency of re-computing full dataset gradient and involves additional gradient evaluations per iteration. 

\begin{algorithm*}[ht]
    \small
    \setstretch{1.3}
    \caption{Black-box variational inference with the joint control variate.}
    \label{alg:final}
    Input step size $\lambda$, negative ELBO estimator $f(w;n,\epsilon)$, and approximation $\tilde f(w;n,\epsilon)$ with closed-form over $\epsilon$.
    
    Initialize parameters $\displaystyle w$ and parameter table $\displaystyle \W=\{\w^1,\ldots,\w^N \}$ using a single epoch with $\naive$.
    
    \parbox{0.30\textwidth}{Initialize running mean.} $\parbox{0.03 \textwidth}{$G$}\leftarrow\E_\m \E_\rxi \nabla \tilde{f}(w;\m,\rxi)$ \hfill $\triangleright$ Sum over $\m$, closed-form over $\rxi$
    
    Repeat until convergence:
    
    \quad Sample $\displaystyle n$ and $\displaystyle \epsilon$.
    
    
     \parbox{0.30\textwidth}{\quad Compute base gradient.} $\displaystyle \parbox{0.03 \textwidth}{$g$} \leftarrow \nabla f(w; n, \epsilon)$
    
     \parbox{0.30\textwidth}{\quad Compute control variate.} $\smash{\displaystyle \parbox{0.03 \textwidth}{$c$} \leftarrow \E_\m \E_\rxi \nabla \tilde{f}(w^\m;\m,\rxi) - \nabla \tilde{f}(\w^\n; n, \epsilon)}$ \hfill $\triangleright$ Use $\smash{\displaystyle \E_\m \E_\rxi \nabla \tilde{f}(w^\m;\m,\rxi)=G}$
    
    \parbox{0.30\textwidth}{\quad Update the running mean.} $\displaystyle \parbox{0.03 \textwidth}{$G$} \leftarrow G + \tfrac{1}{N}\E_\rxi \bigl(\nabla \tilde{f}(w;n,\rxi) - \nabla \tilde{f}(\w^\n;n,\rxi)\bigr)$  \hfill $\triangleright$ Closed-form over $\rxi$
    
    \parbox{0.30\textwidth}{\quad Update the parameter table} $\parbox{0.03 \textwidth}{$w^n$} \leftarrow w$
    
    \parbox{0.30\textwidth}{\quad Update parameters.} $\displaystyle \parbox{0.03 \textwidth}{$w$} \leftarrow w - \lambda (g + c)$ \hfill $\triangleright$ Or use $g+c$ in any stochastic optimization algorithm
\end{algorithm*}

\paragraph{Advantages over existing estimators}
Compared with $\cv$ and $\inc$, $\dual$ can reach arbitrary small gradient variance without lower bound (Eq.~\eqref{eq:dualvarline2}), we empirically verify the lower bounds on two small problems: Fig.~\ref{fig:sonar_aus_grad_var} shows a detailed trace of gradient variance for different estimators using the same optimization trace acquired from $\dual$, where the variance of $\cv$ and $\inc$ both reach the theoretical lower bounds derived in Eq.~\eqref{eq:g_cv_var_final} and Eq.~\eqref{eq:g_inc_var_final}, whereas $\dual$ shows much lower variance. A summarization of variance lower bounds can be seen in Table~\ref{table:estimator_summary}.
Moreover, unlike $\inc$, our proposed $\dual$ estimator controls subsampling noise without the efficiency issue, as $\dual$ only stores (approximate) gradients after integrating over the Monte Carlo variable $\epsilon$, which makes the needed running average independent of $\epsilon$. 

\vspace{-5pt}
\section{RELATED WORK}\label{sec:related}
\vspace{-5pt}

Recently, \citet{boustati2020amortized} proposed to approximate the optimal per-datum control variate for BBVI using a recognition network. This takes subsampling into account. However, like $c_\cv$, this control variate reduces the \emph{conditional} variance of MC noise (conditioned on $n$) but does not address subsampling noise.

Also, \citet{bietti2017stochastic} proposed new incremental gradient method called SMISO, designed for doubly-stochastic problems, which we will compare to below. Intuitively, this uses exponential averages to approximately marginalize out $\epsilon$, and then runs MISO/Finito~\citep{defazio2014finito,mairal2015incremental} (a method similar to SAGA) to reduce subsampling noise. This is similar in spirit to running SGD with a kind of joint control variate. 
However, it is not obvious how to separate the control variate from the algorithm, meaning we cannot use the SMISO idea as a control variate to get a gradient estimator that can be used with other optimizers like Adam, we include a detailed discussion on this issue in Appendix.~\ref{sec:smiso}. Nevertheless, we still include SMISO as one of our baselines.




\vspace{-6pt}
\section{EXPERIMENTS}
\vspace{-6pt}
\label{sec:experiment}

This section evaluates the proposed $\dual$ estimator for BBVI on a range of linear and non-linear probabilistic models, with $208$ to $170K$ samples and latent dimensionalities ranging from $14$ to $85K$. Aside from two toy models (Sonar and Australian) these are large enough that a single full-batch evaluation of $\log p$ takes 15-20 times longer than subsampled valuation, even when implemented on GPU. We compare the proposed $\dual$ estimator against the $\naive$ estimator which controls for no variance, as well as estimators that control for Monte Carlo or data subsampling separately. Our experiments on GPUs show that the $\dual$ estimator's reduced variance leads to better solutions in fewer optimization steps and lower wallclock time. The code can be found at \url{https://github.com/xidulu/JointCV}.

\vspace{-5pt}
\subsection{Experiment setup}
\vspace{-1pt}

\begin{table}
\small
\centering
 \begin{tabular}{llll}
 Task & N & Dims & Model class \tabularnewline
 \hline
 Sonar& 208 & 60 & Logistic regression \tabularnewline
 Australian& 690 & 14 & Logistic regression \tabularnewline
 MNIST& 60,000 & 7,840 & Logistic regression \tabularnewline
 PPCA& 60,000 & 12,544 & Matrix factorization \tabularnewline
 Tennis& 169,405 & 5,525 &  Bradley Terry model \tabularnewline
 MovieLens & 100,000 & 85,050 & Hierarchical model \tabularnewline
\hline 
\end{tabular}
\caption{Dataset size (N), latent dimensionality (Dims) and model class of tasks used in experiments}\label{table:ds_size}
\end{table}
\vspace{-5pt}

\textbf{Tasks and datasets} We evaluate our method by performing BBVI on the following tasks (the complete dataset size and latent dimensionality of each task are provided in table.~\ref{table:ds_size}):
\noindent\begin{itemize}[leftmargin=*, topsep=0pt,parsep=0pt]
\item \textbf{Binary/Multi-class Bayesian logistic regression.} We consider Bayesian logistic regression with standard Gaussian prior for binary classification on the \emph{Sonar} and \emph{Australian} datasets, and multi-class classification on \emph{MNIST}~\citep{lecun1998gradient}.
\item \textbf{Probabilistic principal component analysis (PPCA).} Given a centered dataset $\bx_1, \ldots,\bx_N \in \mathbb{R}^D$, \emph{PPCA}~\citep{tipping1999probabilistic} seeks to extract its principal axes $\bW \in \mathbb{R}^{D \times K}$ assuming
\vspace{-4pt}\[\begin{aligned}
&\bsf{W}_{ij} \sim \mathcal{N}(0, 1), 1 \leq i \leq D, 1 \leq j \leq K, \\
&\rbx_n \sim \mathcal{N}(\bm{0}, \bsf{W} \bsf{W}^\top + \mathrm{diag}(\mathbf{\lambda}^2)).
\end{aligned}\]\vspace{-2pt}
In our experiments, we use BBVI to approximate the posterior over $\bsf{W}$. We test \emph{PPCA} on the standardized training set of MNIST with $K=16$ and $\mathbf{\lambda} = \mathbf{1}$.
\item \textbf{Bradley Terry model for tennis players rating.} Given a set of $N$ tennis match records among $M$ players. Each record has format $(x_{n,1}, x_{n,2}, y_n)$, which denotes a match between players $x_{n,1}$ and $x_{n,2}$ with result $y_n \in \{0,1\}$: $y_n=1$ denotes player $x_{n,1}$ winning the match and vice versa. The Bradley Terry model~\citep{bradley1952rank} assigns each player a score $\theta_m \in \mathbb{R}, m=1,\ldots,M$, and models the match result via
\vspace{-4pt}\[\begin{aligned}
&\rtheta_m \sim \mathcal{N}(0, 1), \\
&\mathsf{y}_n \sim \mathrm{Bernoulli}(\mathrm{logit}^{-1}(\mathsf{\rtheta}_{x_{n,1}} - \mathsf{\rtheta}_{x_{n,2}})).
\end{aligned}\]\vspace{-2pt}
We subsample over matches and perform inference over the score of each player. Following \citet{giordano2024black}, we evaluate the model on men's \emph{tennis} matches log starting from 1960, which contains the results of $169405$ matches among $5525$ players.
\item \textbf{MovieLens analysis with Bayesian hierarchical model.} 
The dataset contains a set of $N$ movie review records from $M$ users, where each record from user $m$ has a feature vector of the movie $\bx_n \in \{0,1\}^{18}$ and a user rating $y_n \in \{1,\ldots,5\}$.
Assigning each user a weight matrix $\bsf{Z}_m \in \mathbb{R}^{18 \times 5}, m=1,\ldots,M$, we model the review through a hierarchical model
\vspace{-4pt}\[\begin{aligned} 
&\bsf{\upmu}_{ij} \sim \mathcal{N}(0, 1), \bsf{\upsigma}_{ij} \sim \mathcal{N}(0, 1), 1 \leq i \leq 18, 1 \leq j \leq 5  \\
&\bsf{Z}_m \sim \mathcal{N}(\bsf{\upmu}, \exp \bsf{\upsigma}),\\
&\mathsf{y}_n  \sim \mathrm{Categorical}\left(\mathrm{softmax}(\bx_n^\top \bsf{Z}_m )\right).
\end{aligned}\vspace{-2pt} 
\]\vspace{-2pt}
We evaluate the model on MovieLens100K~\citep{harper2015movielens}, which has $100,000$ reviews from $943$ users, and perform subsampling over the reviews. 
\end{itemize}




\textbf{Variational distribution.} We focus on mean-field Gaussian BBVI, where the variational distribution follows a factorized Gaussian $\vp = \mathcal{N}(\bmu, \mathrm{diag}(\bsigma^2))$, parameterized by $w=(\bmu, \log \bsigma)$. The mean parameters $\bmu$ is randomly initialized using a standard Gaussian and and we initialize $\log \bsigma$ as $\bm{0}$.

\textbf{Choice of approximation function.} For $\dual$ and $\cv$, we use a second-order Taylor expansion as the approximation function $\tilde{f}(w;n,\epsilon)$ \citep{miller2017reducing}, applied only for the mean parameters $\bmu$, as for mean-field Gaussian BBVI the total gradient variance is often dominated by variance from $\bmu$~\citep{geffner2020approximation}. We provide further details in Appendix.~\ref{appendix:tilde_f_mfvi}.

\textbf{Baselines.} We compare the $\dual$ estimator ($g_\dual$, Eq.~\eqref{eq:g_dual}) with the $\naive$ estimator ($g_\naive$, Eq.~\eqref{eq:g_naive_n_eps}) and the $\cv$ estimator ($g_\cv$, Eq.~\eqref{eq:g_cv_w_n_eps}).
For Sonar and Australian (small datasets) we include the $\inc$ estimator ($g_\inc$, Eq.~\eqref{eq:g_inc_w_n_eps}) as an additional baseline, which requires a full pass through the dataset at each iteration. For larger-scale tasks, the $\inc$ estimator becomes intractable, so we use SMISO instead. 


\textbf{Optimization details.} For the larger-scale MNIST, PPCA, Tennis, and MovieLens, we optimize using Adam~\citep{kingma2014adam}. For the small-scale Sonar and Australian datasets, we use SGD without momentum for transparency. The optimizer for SMISO is pre-determined by its algorithmic structure and cannot be changed. For all estimators, we perform a step-size search to ensure a fair comparison (see Appendix~\ref{sec:step_size_range}),
testing step sizes between $10^{-3}$ and $10^{-1}$ when using Adam and step sizes between $10^{-5}$ and $10^{-2}$ when using SGD. 

\textbf{Mini-batching.} We use mini-batches $\rM$ of data at each iteration (reshuffling each epoch).
For SMISO and the $\inc$ and $\dual$ estimators, we update multiple entries in the parameter table in each iteration and adjust the running mean accordingly. For the Sonar and Australian datasets, due to their small sizes, we use $|\rM|=5$. For all other datasets we use $|\rM|=100$.

\textbf{Evaluation metrics.} We show optimization traces for the best step size chosen retrospectively for each iteration. All ELBO values reported are on the full dataset, estimated with $5000$ Monte Carlo samples. We also show the final ELBO achieved after training vs.\ the step size used to optimize. 
All results reported are averages over multiple independent runs (10 runs for Sonar and Australian datasets, and 5 for the larger scale problems).


\textbf{Experiment environment.} We use JAX~\citep{jax2018github} to implement BBVI, and NumPyro~\citep{phan2019composable} for the models. We conduct all experiments on single GPU machines.



\vspace{-8pt}
\subsection{Results}
\vspace{-2pt}

\begin{figure*}[!ht]
    \centering
    \begin{subfigure}[b]{0.6\linewidth}
        \centering
        \includegraphics[width=\linewidth]{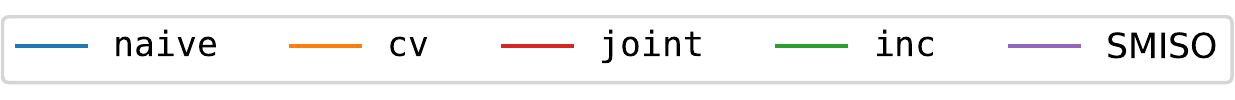}
        \caption*{} 
    \end{subfigure} \\[-4ex]
    \centering
    \begin{subfigure}[b]{\textwidth}
        \centering
        \includegraphics[width=0.48\textwidth]{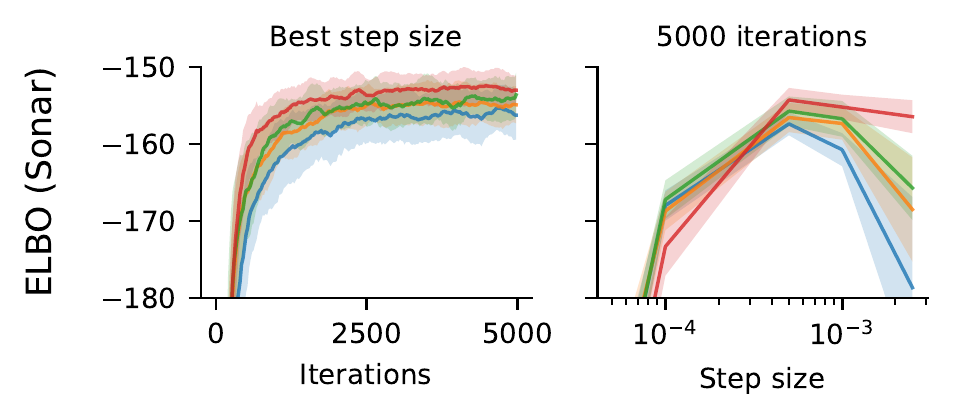}
        \includegraphics[width=0.48\textwidth]{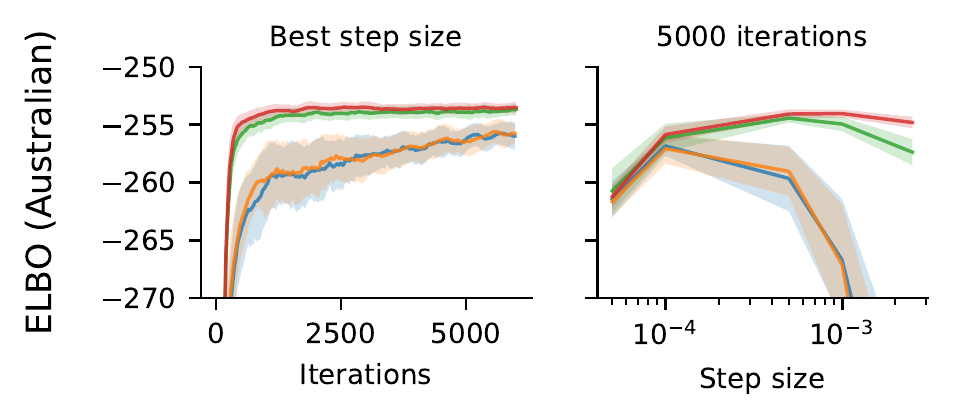}
        \includegraphics[width=0.48\textwidth]{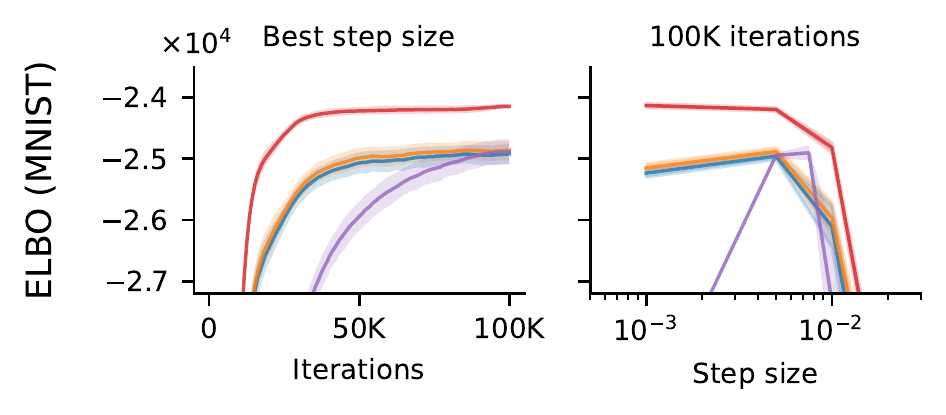}
        \includegraphics[width=0.48\textwidth]{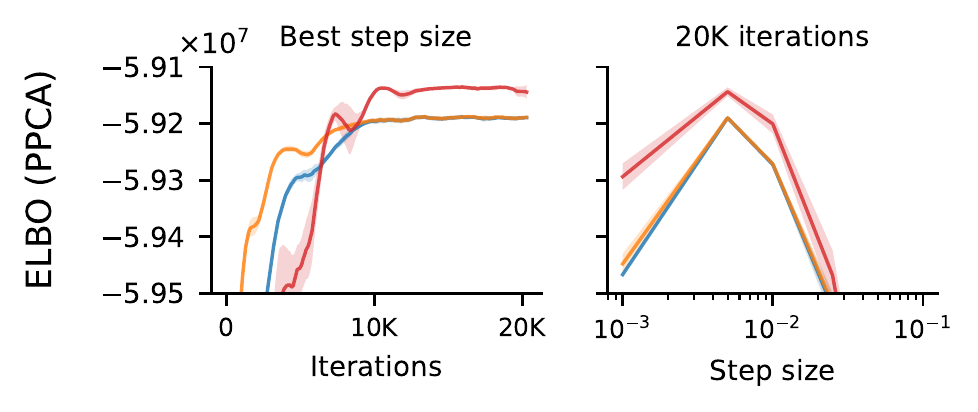}
        \includegraphics[width=0.48\textwidth]{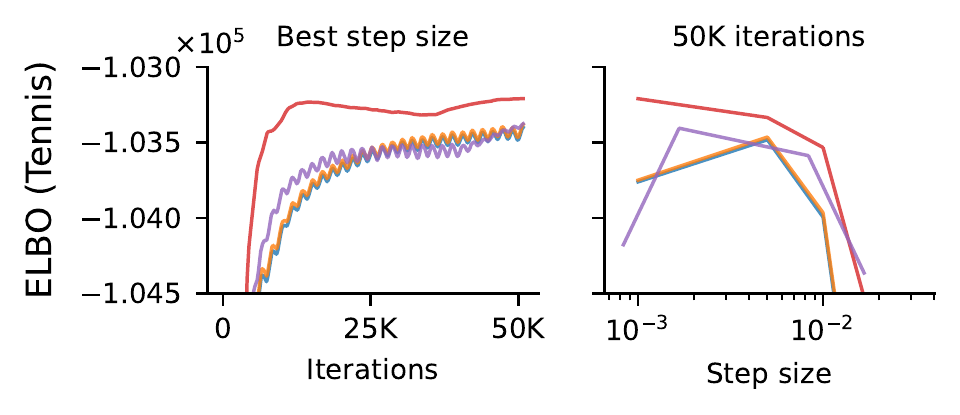}
        \includegraphics[width=0.48\textwidth]{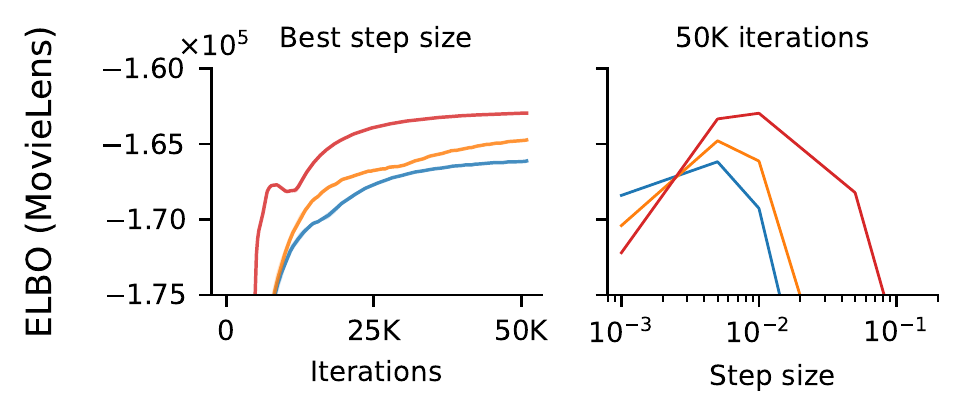}
        \label{fig:best_lr}
    \end{subfigure}
    
    \caption{\textbf{On various tasks, the proposed $\dual$ control variate leads to faster convergence through controlling both Monte Carlo and subsampling noise.} 
    Compared to the $\naive$ estimator, $\cv$ controls only Monte Carlo noise, while  $\inc$ and SMISO control only subsampling noise.
    Our proposed $\dual$ estimator converges faster than $\naive$ and $\cv$ on all tasks. 
    The step sizes for SMISO are rescaled for each model for visualization. On PPCA and MovieLens, SMISO has not converged enough to appear, see Fig.~\ref{fig:sgd_vs_adam} in Appendix.~\ref{sec:sgd_results} for full results. In Tennis, there is periodic behavior for many estimators as gradients have correlated noise that cancels out at the end of each epoch—the $\dual$ estimator largely cancels this. All lines presented the average of multiple trials (5 for Sonar and Australian, 10 for the rest), with shaded areas showing one standard deviation.}
    \label{fig:big_results}
\end{figure*}

On Sonar and Australian, while both the $\inc$ and $\cv$ estimators display lower variance than the $\naive$ estimator, our proposed $\dual$ estimator consistently shows the lowest variance (Fig.~\ref{fig:sonar_aus_grad_var}). This enables the use of larger step sizes, leading to faster convergence (first row in Fig.~\ref{fig:big_results}). 
Notice that, on Austraian, the subsampling noise dominates gradient variance. Thus, $\inc$ shows performance on par with $\dual$. Yet, it is crucial to highlight that $\inc$ requires a full pass over the entire dataset at each optimization step (only possible with small datasets), while $\dual$ does not.
Lastly, we employ MCMC to obtain true posteriors for Sonar and Australian, benefiting from their small scale. The true posterior allows us to measure approximation error by comparing $q_w(z)$'s mean and variance to the true posterior. Results in Fig.~\ref{sec:mcmc_comp} (Appendix.~\ref{sec:mcmc_comp}) confirm that the accelerated convergence from $\dual$ also helps reduce the (mean) approximation error at a faster rate.

The results for large-scale models, MNIST, PPCA, Tennis, and MovieLens, are also presented in Fig.~\ref{fig:big_results} (for these datasets, the $\inc$ estimator is intractable, so we use SMISO as a baseline instead). For MovieLens, as the parameter table required by SAGA does not fit into the GPU memory, we use the SVRG version of the $\dual$ estimator with an update frequency equal to the length of an epoch, introducing one additional gradient call per iteration (amortized).
Broadly, we observe that $\dual$ leads to faster and improved optimization convergence than $\naive$ and $\cv$. 
$\cv$ shows little or no improvement upon $\naive$, which implies that most of the improvement in the $\dual$ estimator comes from reducing subsampling variance.
SMISO, which does not adopt momentum nor adaptive step sizes, suffers from significantly slower convergence, as it requires the use of a considerably smaller step size (to prevent diverging during optimization). We provide comparisons of different estimators using SGD in Appendix.~\ref{sec:sgd_results}.

\begin{table*}
\centering
\scalebox{1.0}{
\small
 \begin{tabular}{lllllll}
 \toprule
 \multirow{2}{*}{Estimator} & \multirow{2}{*}{Variance lower bound} & \multirow{2}{*}{$\nabla f$ evals per iteration} & \multicolumn{4}{c}{Wall-clock time per iteration}\\
\cmidrule{4-7}
  &  &  & MNIST & PPCA & Tennis & MovieLens \tabularnewline
 \midrule
 $\naive$ & $\V_{\n,\repsilon} [\nabla f(w;\n,\repsilon)]$ & 1 & 10.4ms & 12.8ms & 10.2ms & 16.3ms \tabularnewline
 $\cv$ & $\V_{\n} [\nabla f(w;\n)]$ & 2 & 12.8ms & 18.5ms & 14.6ms &  19.6ms \tabularnewline
 $\inc$ & $\V_{\repsilon} [\nabla f(w;\repsilon)]$ & N+2 & 328ms & 897ms & 588ms  & - \tabularnewline
 $\dual$ & $0$ & 3 & 17.6ms & 31.2ms & 29.6ms & 24.4ms \tabularnewline
 Fullbatch-$\naive$ & $\V_{\repsilon} [\nabla f(w;\repsilon)]$ & N & 201ms & 740ms &  203ms & 267ms \tabularnewline
 Fullbatch-$\cv$& $0$ & 2N & 360ms & 1606ms & 246ms & 702ms \tabularnewline
\bottomrule
\end{tabular}}
\caption{Variance, oracle complexity, and wall-clock time for different estimators. Notice that $\inc$ is more expensive than Fullbatch-$\naive$. We hypothesize this is because $\inc$ uses separate $w^n$ for different data points, which is less efficient for parallelism. MovieLens is too large to fit the parameter table into GPU memory, so we use the SVRG version of $\dual$ instead, which requires $4$ gradient evals per iteration (amortized).}
\label{table:estimator_summary}
\end{table*}

\vspace{-5pt}
\subsection{Efficiency analysis}

We now study the computational cost of different estimators. In terms of the number of "oracle" evaluations (i.e.\ evaluations of $f(w;n,\epsilon)$ and its gradient), $\naive$ is the most efficient, requiring a single oracle evaluation per iteration. The $\cv$ estimator requires one gradient and one Hessian-vector product, and the $\dual$ estimator requires one gradient and two Hessian-vector products (one for the control variate and one for updating the running mean $G$.)

Table.~\ref{table:estimator_summary} shows measured measured runtimes
on an Nvidia 2080ti GPU. All numbers are for a single optimization step. For estimators with $\Theta(N)$ oracle complexity (e.g.\ $\inc$) we report average values over 5 steps. For other estimators, we average over 200 steps. Overall, computing the $\dual$ estimator is between 1.5 to 2.5 times slower than computing the $\naive$ estimator, and around 1.2 times slower than $\cv$.
Given that the $\dual$ estimator achieves a given performance using an order of magnitude fewer iterations (Fig.~\ref{fig:big_results}), it leads to significantly faster optimization than the baselines considered. This can be observed in Appendix.~\ref{sec:time}, where we show optimization results in terms of wall-clock time instead of iterations (i.e.\ ELBO vs.\ wall-clock time).




\section*{Acknowledgments}

This material is based upon work supported in part by the National Science Foundation under Grant No. 2045900.

\bibliographystyle{plainnat}
\bibliography{example_paper}

\newpage
\onecolumn
\appendix

\newpage
\section{SMISO}\label{sec:smiso}

In this section, we will have a brief introduction to SMISO~\citep{bietti2017stochastic}. Assume we have a loss function of the form

\begin{equation}
    \E_{\n, \repsilon} f(w ; \n, \repsilon)
\end{equation}

Similar to SAGA~\citep{defazio2014saga}, SMISO maintains a parameter table $W =\{w^1,\ldots,w^N \}$ which stores the parameter value the last time each data point was accessed. SMISO then maintains an average of the value in the parameter table $\bar{w}_k=\E_\n w^\n_k$ where $k$ denotes the $k_{th}$ iteration. $\bar{w}_k$ will later be used as the point for gradient evaluation. Given a randomly drawed sample $n$ and $\epsilon$, SMISO would first update the $n_{th}$ entity in $W$ using exponential average

\begin{equation}
    w_n^{k + 1} = (1 - \alpha) w_n^{k} + \alpha (\bar{w}_{k}  - \gamma \nabla f(\bar{w}_k;\epsilon,n)).
\end{equation}

Then, it updates $\bar{w}_k$ using running average

\begin{equation}\label{eq:smiso_running_mean}
    \bar{w}_{k + 1} = \bar{w}_{k} + \frac{1}{N} w_n^{k + 1} - \frac{1}{N} w_n^{k}.
\end{equation}

If we expand the equation above, we get

\begin{align}
    \bar{w}_{k + 1} &= \bar{w}_{k} + \frac{1}{N} w_n^{k + 1} - \frac{1}{N} w_n^{k} \\
    & = \bar{w}_{k} + \frac{1}{N} \left[  (1 - \alpha) w_n^{k} + \alpha (\bar{w}_{k}  - \gamma \nabla f(\bar{w}_k;\epsilon,n)) - w_n^{k}
    \right] \\
    &= \bar{w}_{k} - \frac{\alpha }{N} \left[\gamma \nabla f(\bar{w}_k;\epsilon,n) + w_n^{k} - \bar{w}_{k}
    \right] \\
    &= \bar{w}_{k} - \frac{\alpha }{N} \left[\gamma \nabla f(\bar{w}_k;\epsilon,n) - (\bar{w}_{k} - w_n^{k})
    \right] 
    \label{eq:smiso_update}
\end{align}

In this case, $\alpha \gamma / N$ is the effective step size. Notice that, if we are using a mini-batch of indices/samples, denoted as $B = \{n_b \}$, in which case multiple entities in the parameter table would be updated in an iteration, then we would have

\begin{align}
    \bar{w}_{k + 1} &= \bar{w}_{k} + \sum_{n_b \in B} \left[\frac{1}{N} w_{n_b}^{k + 1} - \frac{1}{N} w_{n_b}^{k} \right] \\
    &= \bar{w}_{k} - \frac{\alpha |B| }{N} \left[\gamma \E_{\n_b} \nabla f(\bar{w}_k;\epsilon,\n_b) - \E_{\n_b}(\bar{w}_{k} - w_{\n_b}^{k})
    \right] 
\end{align}

in which case the effective step size would become $\frac{\alpha |B| \gamma}{N}$. Therefore, in order to compare SMISO with other estimators using SGD under the same step size, we can first select a range of step sizes for SMISO $\{\gamma_0, \gamma_1, \ldots \}$ and test SGD with step sizes of 
\begin{equation}\label{eq:smiso_stepsize_convertion}
\{\frac{\alpha |B| }{N} \gamma_0, \frac{\alpha |B| }{N} \gamma_1, \ldots \}.
\end{equation}

It is also worth mentioning that, it is not clear to us how to introduce momentum or adaptive step size into SMISO, as we have to strictly follow the running mean update formula (Eq.~\eqref{eq:smiso_running_mean}) to ensure $\E_\n(\bar{w}_{k} - w_\n^{k}) = 0$ for unbiasedness. Adding additional terms (e.g. momentum) or changing the scale of the updates (e.g. normalizing the update by its norm) without careful design could break the unbiasedness. However, studying such modifications is beyond the scope of our paper therefore we only compare our methods with SMISO in its original form.

\section{SVRG version of joint control variate}\label{appendix:SVRG}

\begin{figure}[t]
    \centering
    \includegraphics[width=0.95\linewidth]{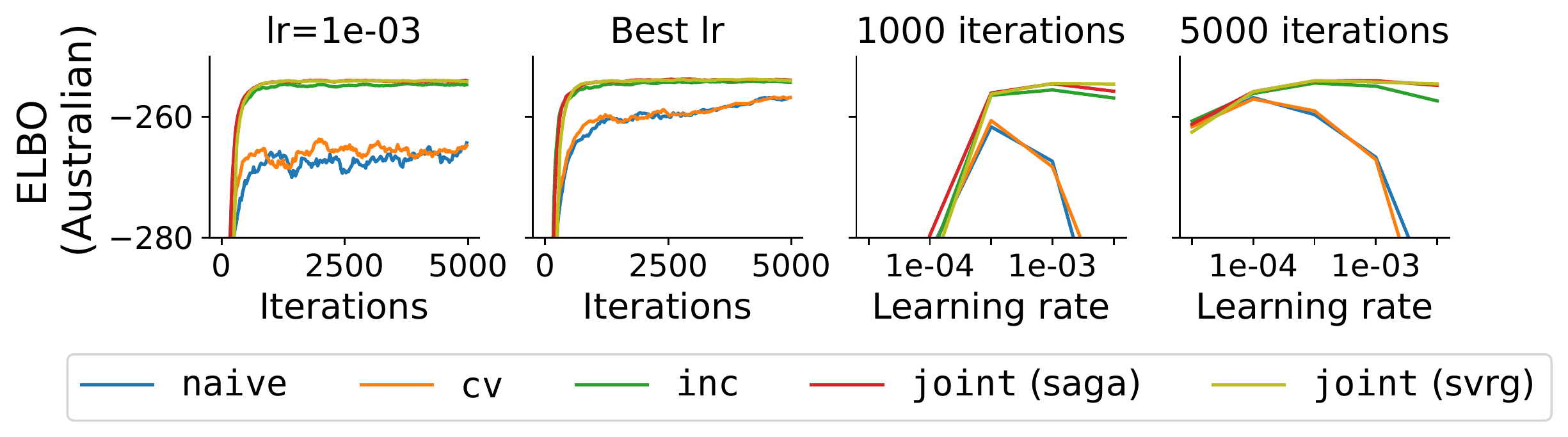}
    \caption{\textbf{The SVRG version of $\dual$ shows performance similar to the SAGA version on Australian.} The origin version of SAGA-based $\dual$ control variate requires $O(ND)$ memory cost. It is possible to alleviate the additional memory cost by using the SVRG version of $\dual$, which costs no extra memory but would require extra gradient evaluation at each step. In the experiments above, we update the SVRG cache every 1 epochs, equivalent to 1 extra gradient evaluation per iteration. Overall, we observe $\dual$ (svrg) showing results similar to the saga version of $\dual$.}
    \label{fig:aus_svrg}
\end{figure}

We present the end-to-end algorithm for applying SVRG version of the $\dual$ control variate in  BBVI in Alg.~\ref{alg:svrg_cv}. On Australian, we find the SVRG version and SAGA version of $\dual$ showing similar performance (Fig.~\ref{fig:aus_svrg}).

\begin{algorithm*}[t]
    \small
    \setstretch{1.3}
    \caption{Black-box variational inference with the joint control variate (SVRG version).}
    \label{alg:svrg_cv}
    Input step size $\lambda$,  negative ELBO estimator $f(w;n,\epsilon)$, and approximation $\tilde f(w;n,\epsilon)$ with closed-form over $\epsilon$.
    
    Input update frequency $K$.
    
    Initialize the parameter $\displaystyle \tilde{w}$.
    
    Repeat until convergence:

    \parbox{0.35\textwidth}{\quad Compute the full gradient of $\tilde{f}$ at $\tilde{w}$.} $\displaystyle \parbox{0.03 \textwidth}{$\bar{g}$} \leftarrow \E_\m \E_\rxi \nabla \tilde{f}(\tilde{w};\m,\rxi)$
    
    \quad Let $w_0 \leftarrow \tilde{w}$.
    
    \quad for $k=1,2,\cdots,K$ do 
    
    \qquad Sample $\displaystyle n$ and $\displaystyle \epsilon$.
    
    
     \parbox{0.30\textwidth}{\qquad Compute base gradient.} $\displaystyle \parbox{0.03 \textwidth}{$g$} \leftarrow \nabla f(w; n, \epsilon)$
    
     \parbox{0.30\textwidth}{\qquad Compute control variate.} $\smash{\displaystyle \parbox{0.03 \textwidth}{$c$} \leftarrow \E_\m \E_\rxi \nabla \tilde{f}(\tilde{w};\m,\rxi) - \nabla \tilde{f}(\tilde{\w}; n, \epsilon)}$ \hfill $\triangleright$ Use $\smash{\displaystyle \E_\m \E_\rxi \nabla \tilde{f}(\tilde{w};\m,\rxi)=\bar{g}}$
    
    \parbox{0.30\textwidth}{\qquad Update parameters.} $\displaystyle \parbox{0.03 \textwidth}{$w_{k}$} \leftarrow w_{k - 1} - \lambda (g + c)$ \hfill $\triangleright$ Or use $g+c$ in any stochastic optimization algorithm

    \quad Update $\tilde{w} \leftarrow w_K$, 
\end{algorithm*}

\section{Derivation of variance for different estimators}

In this section, we will show the full derivation for the trace of the variance of $g_\cv, g_\inc$ and $g_\ens$.

\subsection{Variance of $g_\cv$}\label{sec:cv_grad}


In this section, we will derive the trace for the $\cv$ estimator defined as
\begin{equation}
    g_{\cv}(w; n, \epsilon)= \nabla f(w;n,\epsilon) + \underbrace{\E_\rxi \nabla \tilde{f}(w;n, \rxi) - \nabla \tilde{f}(w; n, \epsilon)}_{c_{\cv}(w;n, \epsilon)},
\end{equation}
where $\tilde{f}$ is an approximation function of $f$ with closed-form expectation with respect to $\epsilon$.

To start with, we will apply the law of total variance
\begin{align}
\V[g_{\cv}] &= \E_{\n}\V_{\repsilon} g_{\cv} + \V_{\n}\E_{\repsilon} g_\cv.
\end{align}
The first term can be computed as
\begin{align}
  \E_{\n}\V_{\repsilon} g_{\cv} &=  \E_{\n}\V_{\repsilon}[\nabla f(w;\n,\repsilon) + \E_\rxi \nabla \tilde{f}(w;\n,\rxi) - \nabla \tilde{f}(w; \n, \repsilon)] \\
  &= \E_{\n}\V_{\repsilon}[\nabla f(w;\n,\repsilon) -\nabla \tilde{f}(w; \n, \repsilon)],
\end{align}
which follows since $\E_\rxi \nabla \tilde{f}(w;\n,\rxi)$ is a constant with respect to $\epsilon$ and therefore does not affect the variance.

The second term can be computed as
\begin{align}
    \V_{\n}\E_{\repsilon} g_\cv &= \V_{\n}\E_{\repsilon}[\nabla f(w;\n,\repsilon) + \E_\rxi \nabla \tilde{f}(w;\n, \rxi) - \nabla \tilde{f}(w; \n, \repsilon)]  \\
    &= \V_{\n}\left[\E_{\repsilon}[\nabla f(w;\n,\repsilon)] + \E_{\repsilon}[\E_\rxi \nabla \tilde{f}(w;\n, \rxi)] - \E_{\repsilon}[\nabla \tilde{f}(w; \n, \repsilon)] \right] \\
    &= \V_{\n}\E_{\repsilon}[\nabla f(w;\n,\repsilon) + \nabla \tilde{f}(w;\n) - \nabla \tilde{f}(w; \n)] \\
    &= \V_{\n}\E_{\repsilon}[\nabla f(w;\n,\repsilon)] \\
    &=  \V_\n[\nabla f(w;\n)].
\end{align}

Then we can combine the two terms together to get
\begin{equation}
\V[g_{\cv}]= \E_{\n}\V_{\repsilon}[\nabla f(w;n,\epsilon) -\nabla \tilde{f}(w; \n, \repsilon)]+
\V_\n[\nabla f(w;\n)]
\end{equation}

\subsection{Variance of $g_\inc$} \label{sec:inc_grad}
Here, we will derive the trace of the variance of the $\inc$ estimator defined as
\begin{equation}
    g_{\inc}(w;n, \epsilon) = \nabla f_n(w;n, \epsilon) + \underbrace{\E_{\m} \nabla f(w^{\m};\m, \epsilon) - \nabla f (w^{n};n, \epsilon)}_{c_\inc(w;n,\epsilon)}.
\end{equation}

We can derive its variance by first applying the law of total variance
\begin{align}
\V[g_{\inc}] &= \E_{\repsilon}\V_{\n}g_{\inc} + \V_{\repsilon}\E_{\n}g_\inc .
\end{align}

The first term can be computed as
\begin{align}
  \E_{\repsilon}\V_{\n} g_{\inc} &= \E_{\repsilon}\V_{\n}[\nabla f_n(w;\n, \repsilon) + \E_{\m} \nabla f(w^{\m};\m, \epsilon) - \nabla f (w^{\n};\n, \repsilon)] \\
  &= \E_{\repsilon}\V_{\n}[\nabla f(w;\n,\repsilon) -\nabla f (w^{\n};\n, \repsilon)],
\end{align}
where the second line follows because $\E_{\m}\nabla f(w^{\m};\m, \epsilon)$ is a constant with respect to $n$. 

The second term can be computed as
\begin{align}
    \V_{\repsilon}\E_{\n} g_\inc &= \V_{\repsilon}\E_{\n}[\nabla f_n(w;\n, \repsilon) + \E_{\m}\nabla f(w^{\m};\m, \epsilon) - \nabla f (w^{\n};\n, \repsilon)]  \\
    &= \V_{\repsilon}\left[ \E_{\n}\nabla f_n(w;\n, \repsilon) + \E_{\n}\E_{\m}\nabla f(w^{\m};\m, \epsilon) - \E_{\n}\nabla f (w^{\n};\n, \repsilon) \right] \\
    &=\V_{\repsilon}\left[ \E_{\n}[\nabla f_n(w;\n, \repsilon)] +  \E_{\m} \nabla f(w^{\m};\m, \epsilon) - \E_{\n}\nabla f (w^{\n};\n, \repsilon)\right] \\
    &= \V_{\repsilon}\E_{\n}[\nabla f_n(w;\n, \repsilon)] \\
    &=  \V_\repsilon[\nabla f(w;\repsilon)],
\end{align}
which then leads us to
\begin{equation}
\V[g_{\inc}]= \E_{\repsilon}\V_{\n}[\nabla f(w;\n,\repsilon) -\nabla f (w^{\n};\n, \repsilon)]+
\V_{\repsilon}[\nabla f(w;\repsilon)].
\end{equation}

\subsection{Variance of $g_\ens$}\label{sec:combo_grad}


In this section, we will derive the variance for the estimator $g_\ens$ defined as
\begin{equation}
g_\ens(w;n,\epsilon) = \nabla f(w; n, \epsilon) + \underbrace{\beta  c_{\cv}(w;n, \epsilon) + (1 - \beta) c_{\inc}(w;n, \epsilon)}_{c_{\ens}(w;n,\epsilon)},
\end{equation}
under the ideal assumption where we have $f=\tilde{f}$ and $w=w^n, \forall n$. The variance can be derived through
\begin{align}
    \V[g_\ens] &=\V_{\repsilon,\n}[\nabla f(w; \n, \repsilon) + \beta  c_{\cv}(n, \repsilon) + (1 - \beta) c_{\inc}(\n, \repsilon)] \\
    &= \V_{\repsilon,\n}\bigg[\nabla f(w; \n, \repsilon) + \beta \left(\E_\rxi \nabla \tilde{f}(w;\n, \rxi) - \nabla \tilde{f}(w; \n, \repsilon) \right) + \\
   &\qquad\qquad\qquad\qquad (1 - \beta) \left( \nabla \E_{\m} f(w^{\m};\m, \repsilon) - \nabla f (w^{\n};\n, \repsilon) \right)\bigg] \nonumber \\
   \shortintertext{Then we replace $\tilde{f}$ with $f$ and $w^n$ with $w$ based on our assumption, }
    \V[g_\ens]   &= \V_{\repsilon,\n}\bigg[\nabla f(w; \n, \repsilon) + \beta \left(\E_\rxi \nabla {f}(w;\n, \rxi) - \nabla {f}(w; \n, \repsilon) \right) + \\
   &\qquad\qquad\qquad\qquad (1 - \beta) \left( \nabla \E_{\m} f(w;\m, \repsilon) - \nabla f (w;\n, \repsilon) \right)\bigg] \nonumber \\
    &= \V_{\repsilon,\n}\bigg[\nabla f(w; \n, \repsilon) + \beta\left( \nabla {f}(w;\n) - \nabla f(w; \n, \repsilon)\right) + (1-\beta)\left(f(w;\repsilon) - f(w;\n,\repsilon) \right)   \bigg] \\
    &=\V_{\repsilon,\n}\Bigg[\beta \nabla f(w;\n) + (1-\beta)\nabla f(w;\repsilon)  \Bigg] \\
    &= \beta^2 \V_{\n}[\nabla{f}(w;\n)] + (1-\beta)^2 \V_{\repsilon}[\nabla f(w;\repsilon)].
\end{align}
The last line follows because $\nabla{f}(w;n)$ is independent of $\nabla{f}(w;\epsilon)$.

\section{Step-size search range}
\label{sec:step_size_range}






For Australian and Sonar, we experiment with learning rates of
$$\{7.5 \times 10^{-3}, 5 \times 10^{-3}, 2.5 \times 10^{-3}, 1 \times 10^{-3}, 5 \times 10^{-4}, 1 \times 10^{-4}, 5 \times 10^{-5}, 2.5 \times 10^{-5}, 1 \times 10^{-5}\}$$

For MNIST, PPCA ,Tennis and MovieLens, we used 
$$\{ 1 \times 10^{-1}, 5 \times 10^{-2}, 1 \times 10^{-2}, 5 \times 10^{-3}, 1 \times 10^{-3} \}$$ for $\naive$, $\cv$ and $\dual$, where the optimizer is Adam.

When optimizing with SMISO, we set $\alpha=0.9$ and we perform grid search over the value of $\gamma$, for MNIST with SMISO, we experiment with $\gamma$ in $$\{5 \times 10^{-2}, 2.5 \times 10^{-2}, 1 \times 10^{-2}, 5 \times 10^{-3}, 2.5 \times 10^{-3}, 1 \times 10^{-3}, 5 \times 10^{-4}, 1 \times 10^{-4}, 5 \times 10^{-5}, 1 \times 10^{-5}\}$$

For Tennis with SMISO, we experiment with $\gamma$ in $$
\{ 5 \times 10^{-2}, 2.5 \times 10^{-2}, 1 \times 10^{-2}, 5 \times 10^{-3}, 1 \times 10^{-3}, 1 \times 10^{-4}, 1 \times 10^{-5}\}
$$

For PPCA with SMISO, we experiment with $\gamma$ in $$\{1 \times 10^{-2}, 5 \times 10^{-3}, 1 \times 10^{-3}, 1 \times 10^{-4}, 1 \times 10^{-5}, 1 \times 10^{-6}, 1 \times 10^{-7}\}$$

For MovieLens with SMISO, we experiment with $\gamma$ in $$\{2.5 \times 10^{-3}, 1 \times 10^{-3}, 5 \times 10^{-4}, 1 \times 10^{-4}\}$$

\newpage
\section{Generic optimization algorithm}
\label{sec:generic_opt}

\begin{algorithm*}[t]
    \small
    \setstretch{1.3}
    \caption{Joint control variate for generic doubly-stochastic optimization problem.}
    \label{alg:generic}
    Input step size $\lambda$, doubly-stochastic objective $f(w;n,\epsilon)$, and approximation $\tilde f(w;n,\epsilon)$ with closed-form over $\epsilon$.
    
    Initialize parameters $\displaystyle w$ and parameter table $\displaystyle \W=\{\w^1,\ldots,\w^N \}$ using a single epoch with $\naive$.
    
    \parbox{0.30\textwidth}{Initialize running mean.} $\parbox{0.03 \textwidth}{$G$}\leftarrow\E_\m \E_\rxi \nabla \tilde{f}(w;\m,\rxi)$ \hfill $\triangleright$ Sum over $\m$, closed-form over $\rxi$
    
    Repeat until convergence:
    
    \quad Sample $\displaystyle n$ and $\displaystyle \epsilon$.
    
    
     \parbox{0.30\textwidth}{\quad Compute base gradient.} $\displaystyle \parbox{0.03 \textwidth}{$g$} \leftarrow \nabla f(w; n, \epsilon)$
    
     \parbox{0.30\textwidth}{\quad Compute control variate.} $\smash{\displaystyle \parbox{0.03 \textwidth}{$c$} \leftarrow \E_\m \E_\rxi \nabla \tilde{f}(w^\m;\m,\rxi) - \nabla \tilde{f}(\w^\n; n, \epsilon)}$ \hfill $\triangleright$ Use $\smash{\displaystyle \E_\m \E_\rxi \nabla \tilde{f}(w^\m;\m,\rxi)=G}$
    
    \parbox{0.30\textwidth}{\quad Update the running mean.} $\displaystyle \parbox{0.03 \textwidth}{$G$} \leftarrow G + \tfrac{1}{N}\E_\rxi \bigl(\nabla \tilde{f}(w;n,\rxi) - \nabla \tilde{f}(\w^\n;n,\rxi)\bigr)$  \hfill $\triangleright$ Closed-form over $\rxi$
    
    \parbox{0.30\textwidth}{\quad Update the parameter table} $\parbox{0.03 \textwidth}{$w^n$} \leftarrow w$
    
    \parbox{0.30\textwidth}{\quad Update parameters.} $\displaystyle \parbox{0.03 \textwidth}{$w$} \leftarrow w - \lambda (g + c)$ \hfill $\triangleright$ Or use $g+c$ in any stochastic optimization algorithm
\end{algorithm*}

In Alg.~\ref{alg:final}, we describe the end-to-end procedure of applying joint control variate in BBVI. The joint control variate can also be applied in generic doubly-stochastic optimization problems as is shown in Alg.~\ref{alg:generic}.

We evaluate the generic version on generalized linear models with Gaussian dropout, with an objective function defined as
\begin{align}
    &f^{\lr}(w) = \E_{\n}\E_{\repsilon}f^{\lr}(w;\n,\repsilon), \label{eq:glm_obj} \\ 
    &f^{\lr}(w;n,\epsilon):= \mathcal{L}\left(y_{n}, \phi(x_n;w,\epsilon)\right)\label{eq:f_lr_w_n_eps} \\
    &\phi(x_n;w,\epsilon)=w (\epsilon \odot x_{n}),
\end{align}
where $x_n \in \mathbb{R}^{D}, y_n \in \mathbb{R}^{K}, w \in \mathbb{R}^{K \times D}$ and $\epsilon \in \mathbb{R}^{D}$ is a sample from $\mathcal{N}(\bm{1}, \sigma \mathbb{I})$, $\odot$ stands for element-wise product and $\mathcal{L}$ is a loss function such as mean-squared error.

We can find an approximation to Eq.~\eqref{eq:f_lr_w_n_eps} by applying second-order Taylor expansion around $\epsilon=1$, given by
\begin{equation}\label{eq:f_glm_approx}
    \tilde{f}^{\lr}(w;n,\epsilon) = f^{\lr}(w;n,1)  +  (\epsilon - 1)^\top \nabla_\epsilon f^{\lr}(w;n,1) 
    +  \frac{1}{2}(\epsilon - 1)^\top \nabla^2_{\epsilon}f^{\lr}(w;n,1)(\epsilon - 1),
\end{equation}
whose expectation with respect to $\epsilon$ can be given in closed-form as
\begin{equation}\label{eq:e_f_glm_approx}
    \E_{\repsilon}\tilde{f}^{\lr}(w;n,\repsilon) = f^{\lr}(w;n,1) + 
    \frac{\sigma^2}{2} \mathrm{tr}\left(\nabla^2_{\epsilon}f^{\lr}(w;n,1)\right).
\end{equation}

\begin{figure}[h]
\centering
\includegraphics[width=0.9\textwidth]{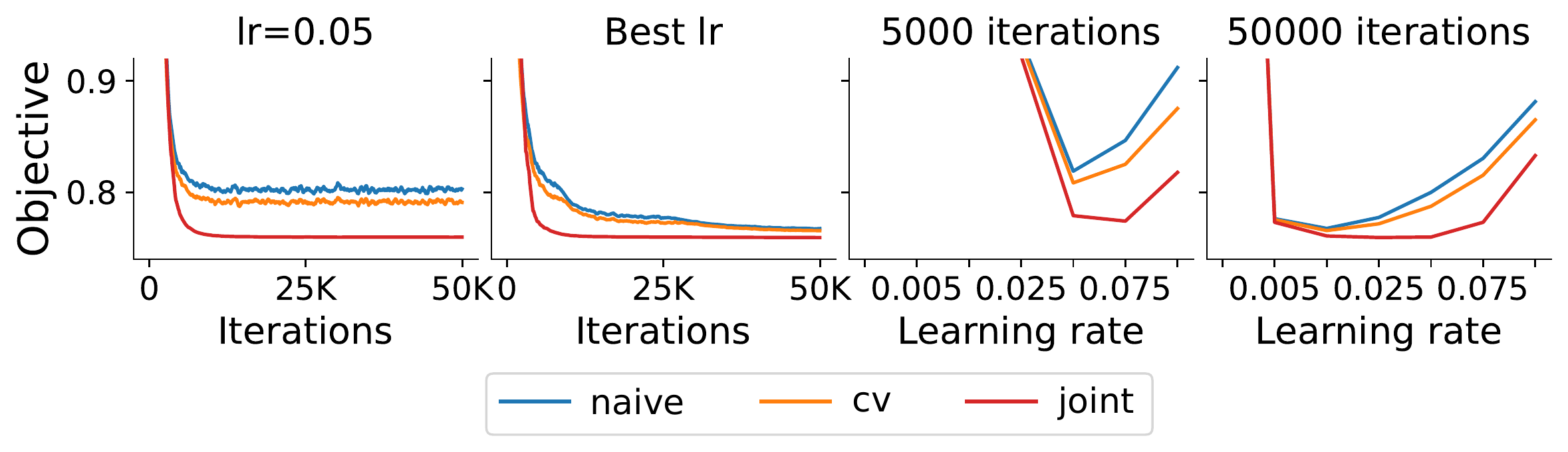}
\caption{\textbf{The joint estimator leads to improved convergence at higher learning rates on Gaussian dropout on CIFAR-10.} (For small enough learning rates, optimization speed is limited by the learning rate itself and so all estimators perform identically.) The first column shows the trace of the objective (logistic loss) under a learning rate of $0.05$. The second column shows the trace of the objective under the best learning rate chosen retrospectively at each iteration. The final two columns show the objective as a function of different learning rates at two different numbers of iterations. Note that the learning that the $\dual$ has its best performance at a higher learning rate than the other estimators. (the $\inc$ estimator is too expensive to be included here.)}
\label{fig:cifar_results}
\end{figure}

\paragraph{Results} We compare the performance of $g_{\naive}^{\lr}$, $g_{\cv}^{\lr}$, and $g_{\dual}^{\lr}$ on CIFAR-10~\citep{krizhevsky2009learning} classification, where we apply dropout on features extracted from a LeNet~\citep{lecun1998gradient} pretrained on CIFAR-10 and then fine-tune the output layer using the cross-entropy loss with $\sigma=0.5$. We use a batch size of 100, and optimize using standard gradient descent without momentum for a wide range of learning rates. We present the results in Figure~\ref{fig:cifar_results} where we show the trace of objective evaluated on the full training set under different learning rates and different numbers of iterations. We can see that $g_\dual$ always reaches objectives smaller than the baseline estimators, displaying significantly better convergence for large learning rates.

\section{Approximation function for mean-field Gaussian BBVI}\label{appendix:tilde_f_mfvi}

Recall that, given $w=(\bmu, \log \bsigma)$, the objective function for mean-field Gaussian BBVI is written as 
\begin{align}
    f(w)^{\vi} &= \E_\n\E_{\repsilon} f(w; \n, \epsilon), \\
    \text{where}\ f(w; n, \epsilon) &= -N \log p(x_n \mid \mathcal{T}_w(\epsilon) ) - \log p(\mathcal{T}_w(\epsilon))- \mathbb{H}(w), \quad \mathcal{T}_w(\epsilon) = \bmu + \epsilon \odot \bsigma,
\end{align}
where we use the $\epsilon$ notation here to also represent a vector.
Inspired by previous work \citep{miller2017reducing}, we get an approximation for $f^{\vi}(w;n,\epsilon)$ using a second order Taylor expansion for the negative total likelihood  $k_n(z) = N \log p(x_n \mid z) + \log p(z)$ around $z_0 = \mathcal{T}_{w}(0)$\footnote{We use $z_0=\mathrm{stop\_gradient}\left(\mathcal{T}_{w}(0)\right)$ so that the gradient does not backpropagate from $z_0$ to $w$.}, which yields
\begin{equation}\label{eq:bbvi_taylor_approx}\small
    \tilde{f}^{\vi} (w;n,\epsilon) = k_{n}(z_0) + (\tw - z_0)^\top \nabla k_{n}(z_0) +
    \frac{1}{2} (\tw - z_0)^\top \nabla^2 k_{n}(z_0) (\tw - z_0) + \mathbb{H}(w),
\end{equation}
where we assume the entropy can be computed in closed form. The approximation function's gradient with respect to the variational parameter is given by:
\begin{align}
    \nabla_{w} {\tilde{f}(w;n,\epsilon)} &= \left[\frac{\partial\left( \tw - z_0\right)}{\partial w}\right]^\top\left[ \nabla k_{n}(z_0) + \nabla^2 k_{n}(z_0) (\bmu + \epsilon \odot \bsigma - \bmu + \bm{0} \odot \bsigma) \right] + \nabla_w \mathbb{H}(w),\\
    &= \left[\frac{\partial\left( \tw \right)}{\partial w}\right]^\top \left[ \nabla k_{n}(z_0) + \nabla^2 k_{n}(z_0) (\epsilon \odot \bsigma) \right] + \nabla_w \mathbb{H}(w),
\end{align}
where $\frac{\partial\left( \tw \right)}{\partial w}$ denotes Jacobian matrix. Note that, despite the gradient computation involving the Hessian, it can be computed efficiently without explicitly storing the Hessian matrix through Hessian vector product. However, the \emph{expectation} of the gradient can only be can only be computed efficiently with respect to the mean parameter $\bmu$ but not for the scale parameter $\bsigma$. To see that, we first compute the expected gradient with respect to $\bmu$, using the fact that $\frac{\partial\left( \tw - z_0\right)}{\partial \bmu} = \bm{I}$ and $\epsilon$ is zero-mean:
\begin{equation}
    \E_\repsilon \nabla_{\bmu} {\tilde{f}(w;n,\repsilon)} = \nabla  k_{n}(z_0) + \nabla_{\bmu} \mathbb{H}(w)
\end{equation}
The expected gradient with respect to $\bsigma$ is given by:
\begin{align}
    \E_\repsilon \nabla_{\bsigma} {\tilde{f}(w;n,\repsilon)} &= \E_\repsilon{\left[\diag (\repsilon) \left[ \nabla k_{n}(z_0) + \nabla^2 k_{n}(z_0) (\repsilon \odot \bsigma) \right] \right]} + \nabla_{\bsigma} \mathbb{H}(w),\\
    &= \E_\repsilon{\left[\repsilon \odot \nabla k_{n}(z_0) + \nabla^2 k_{n}(z_0) (\repsilon^2 \odot \bsigma)\right]} + \nabla_{\bsigma} \mathbb{H}(w),  \\
    &= \diag\left(\nabla^2 k_{n}(z_0)\right) \odot \bsigma + \nabla_{\bsigma} \mathbb{H}(w),
\end{align}
which requires the diagonal of the Hessian, causing computing difficulty in many problems. This means $g_{\cv}^{\vi}(w;n, \epsilon)$ and $g_{\dual}^{\vi}(w;n, \epsilon)$ can only be efficiently used as the gradient estimator for $\bmu$. Fortunately, controlling only the gradient variance on $\bmu$ often means controlling most of the variance, as, with mean-field Gaussians, the total gradient variance is often dominated by variance from $\bmu$ \citep{geffner2020approximation}.

\section{Comparision with true posterior}\label{sec:mcmc_comp}

On small problems, in particular, Sonar and Australian, we can acquire the true posterior using MCMC and compute the approximation error of the variational posterior using the true posterior. To be more specific, we run MCMC on these two problems using NUTS~\citep{hoffman2014no} with 4 chains, where we warm up for 5,000 steps and then collect 25,000 samples from each chain, giving a total of 100K samples that we use to estimate the mean and variance of the true posterior, denoted as $\bmu_{\rm mcmc}$ and $\bsigma^2_{\rm mcmc}$ respectively. We then compute the L2 distance between the mean and variance of the variational posterior (a diagonal Gaussian) and that of the true posterior as the approximation error. We additionally acquire a set of ground truth variational parameters using full dataset and 200 Monte Carlo samples for gradient estimation, optimized for 10,000 iterations with a learning rate of $1\times 10^{-4}$. The approximation error based on the ground truth parameter serves as a reference value on the \emph{smallest} error each estimator can achieve.

\begin{figure}
    \centering
\begin{subfigure}{0.9\textwidth}
    \centering
       \includegraphics[width=0.8\textwidth]{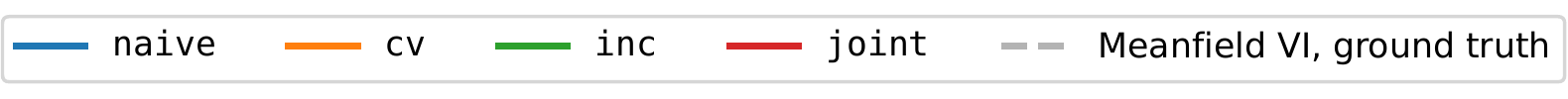}
\end{subfigure}
\begin{subfigure}{0.9\textwidth}
    \centering
       \includegraphics[width=\textwidth]{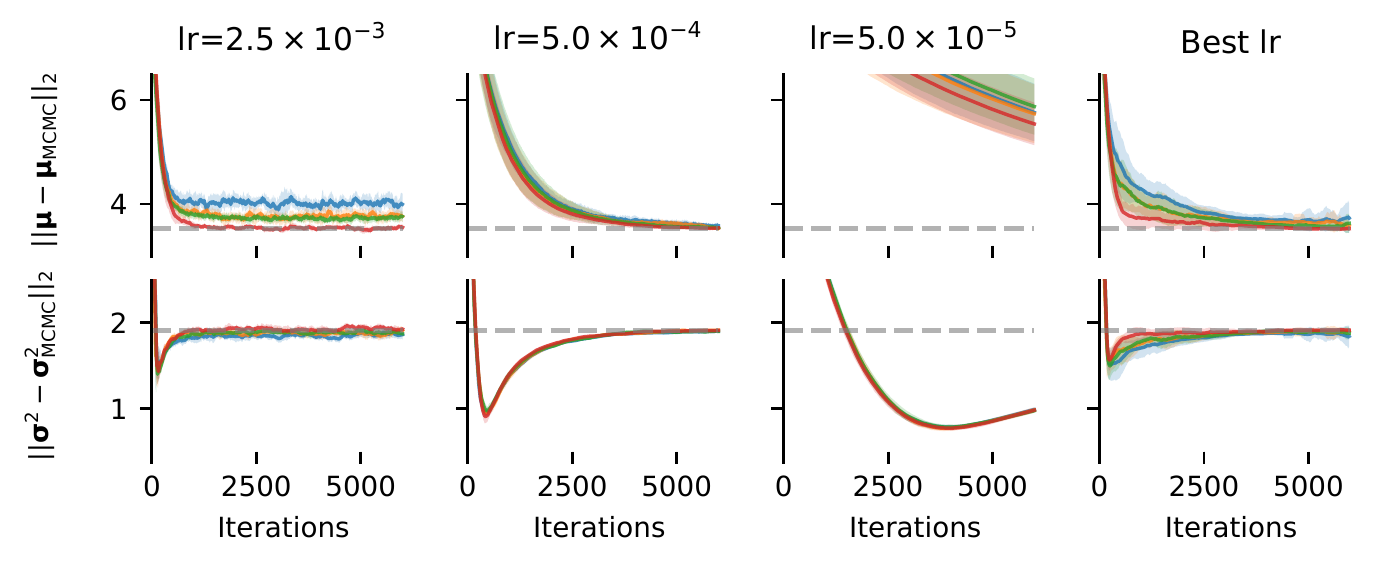}
    \caption{Sonar}
\end{subfigure}
\begin{subfigure}{0.9\textwidth}
    \centering
       \includegraphics[width=\textwidth]{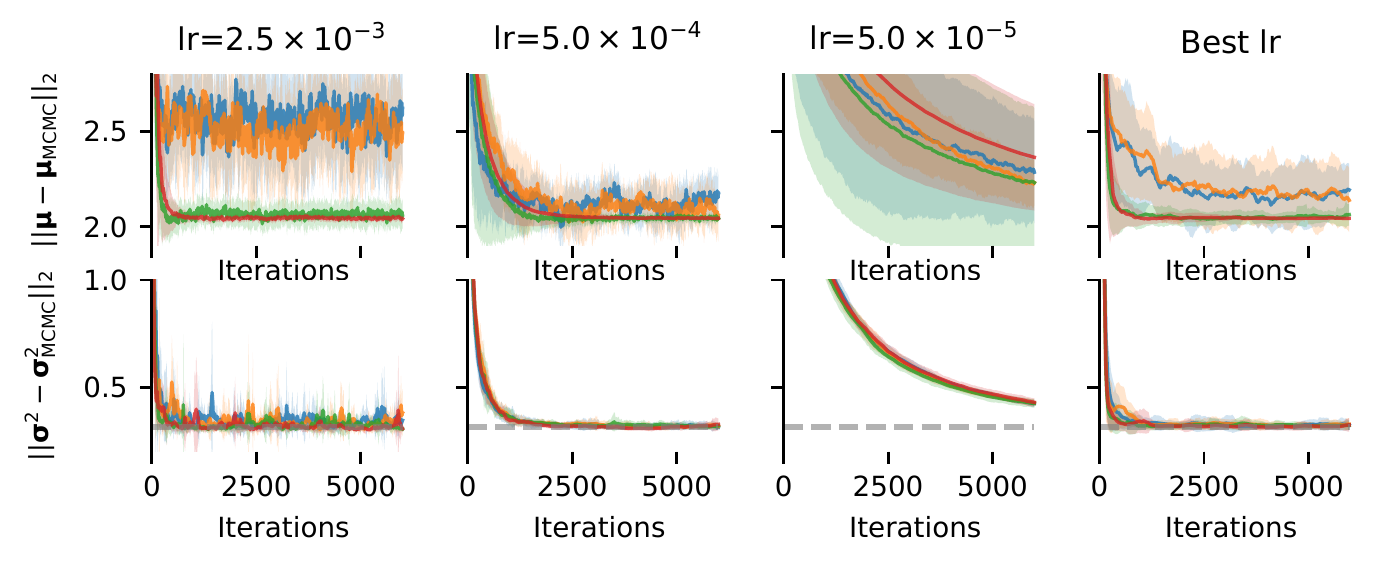}
       \caption{Australian}
\end{subfigure}
    \caption{\textbf{Meanfield VI under the proposed $\dual$ estimator approaches true posterior mean faster}. On small-scale problems, we compare the mean and variance of the variational posterior (a diagonal Gaussian) to that of the true posterior estimated with MCMC using 100K samples, with approximation error measured by L2 distance. The last column shows the error trace under the best learning rate chosen retrospectively at each iteration based on the ELBO. The grey dashed lines show the error under ground truth variational parameters acquired with full batch gradient and 200 Monte Carlo samples per iteration, representing the best error each estimator can achieve.
    On the mean parameter, $\dual$ reduces the error faster than baseline estimators. 
    For variance errors, all estimators demonstrate similar behavior, as we are only controlling the variance on the mean parameters. The results presented are based on 10 random trials, where the solid lines denote the averaged values and the shaded area represents the one standard deviation.
    }
    \label{fig:mcmc_comparision}
\end{figure}

The results are presented in Fig.~\ref{fig:mcmc_comparision}, the observation aligns with the ELBO traces (Fig.~\ref{fig:big_results}), where $\dual$ is capable of approaching the true posterior mean at a speed faster than with baseline estimators, eventually reach the approximation error of the ground truth variational parameters.

\section{Results under SGD}\label{sec:sgd_results}

\begin{figure}[ht]
    \centering
\begin{subfigure}{0.485\textwidth}
    \centering
       \includegraphics[width=\textwidth]{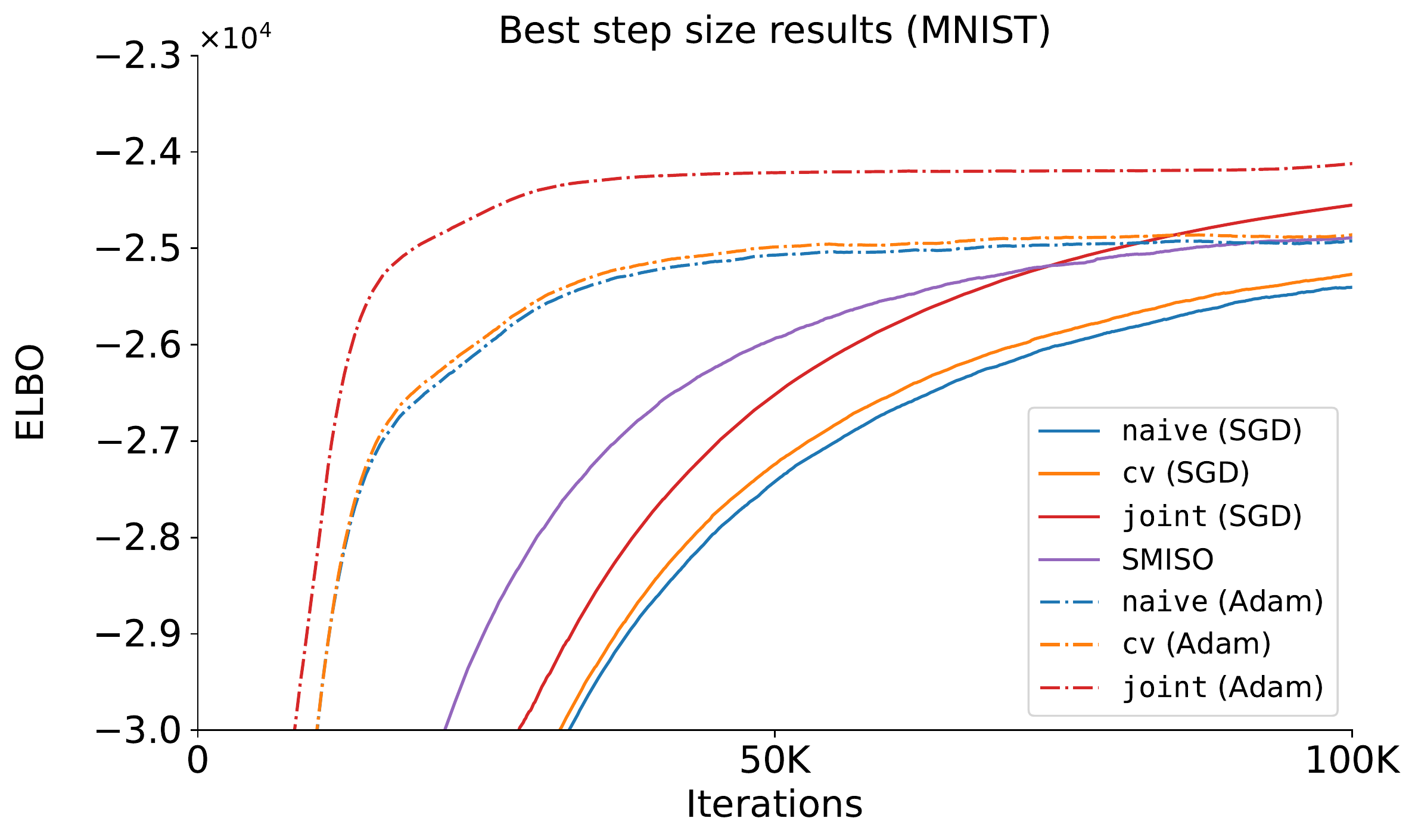}
  \label{fig:mnist_sgd_adam}   
\end{subfigure}
    \begin{subfigure}{0.485\textwidth}
        \centering
        \includegraphics[width=\textwidth]{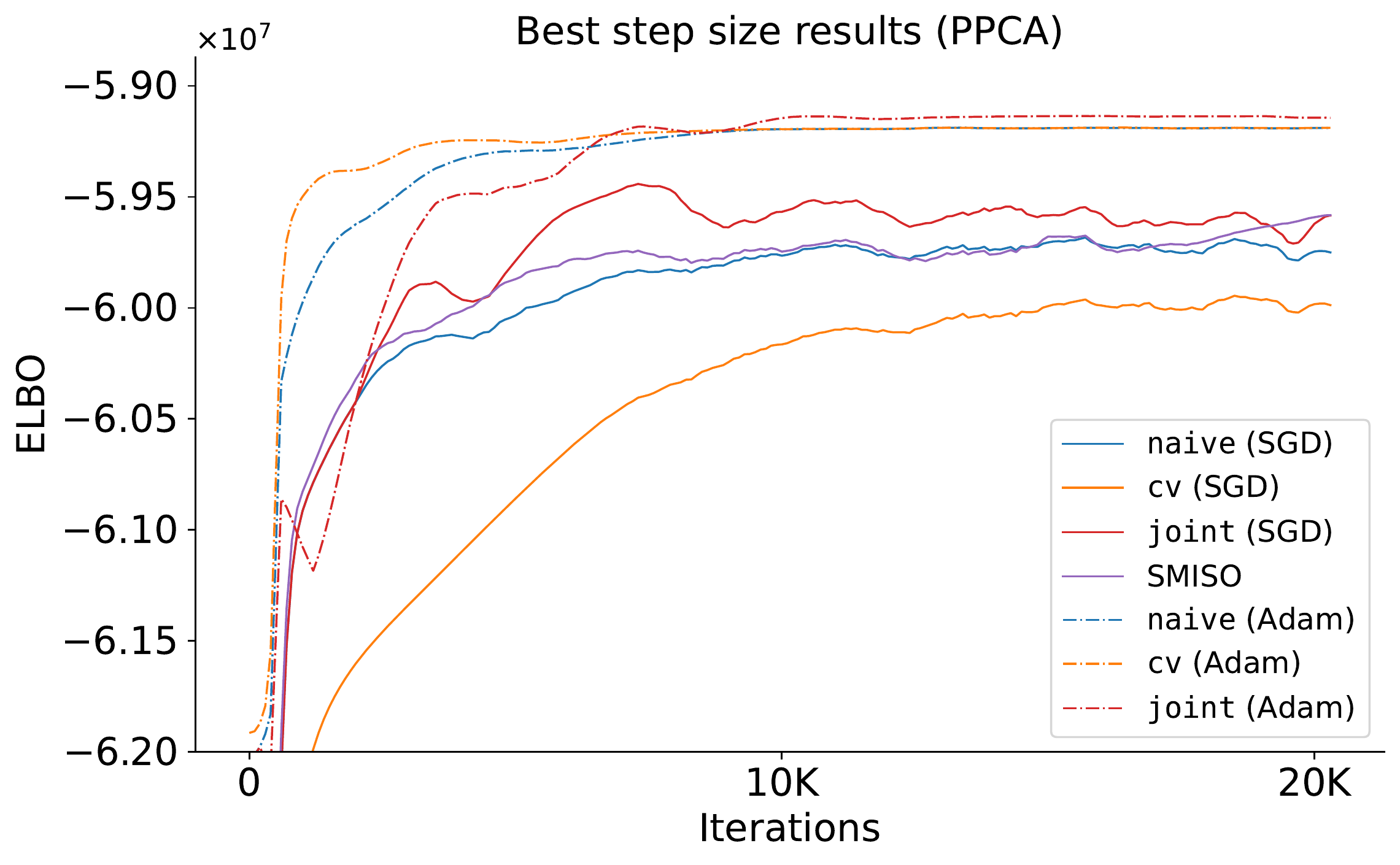}
        \label{fig:ppca_sgd_adam}
    \end{subfigure}
\begin{subfigure}{0.485\textwidth}
    \centering
       \includegraphics[width=\textwidth]{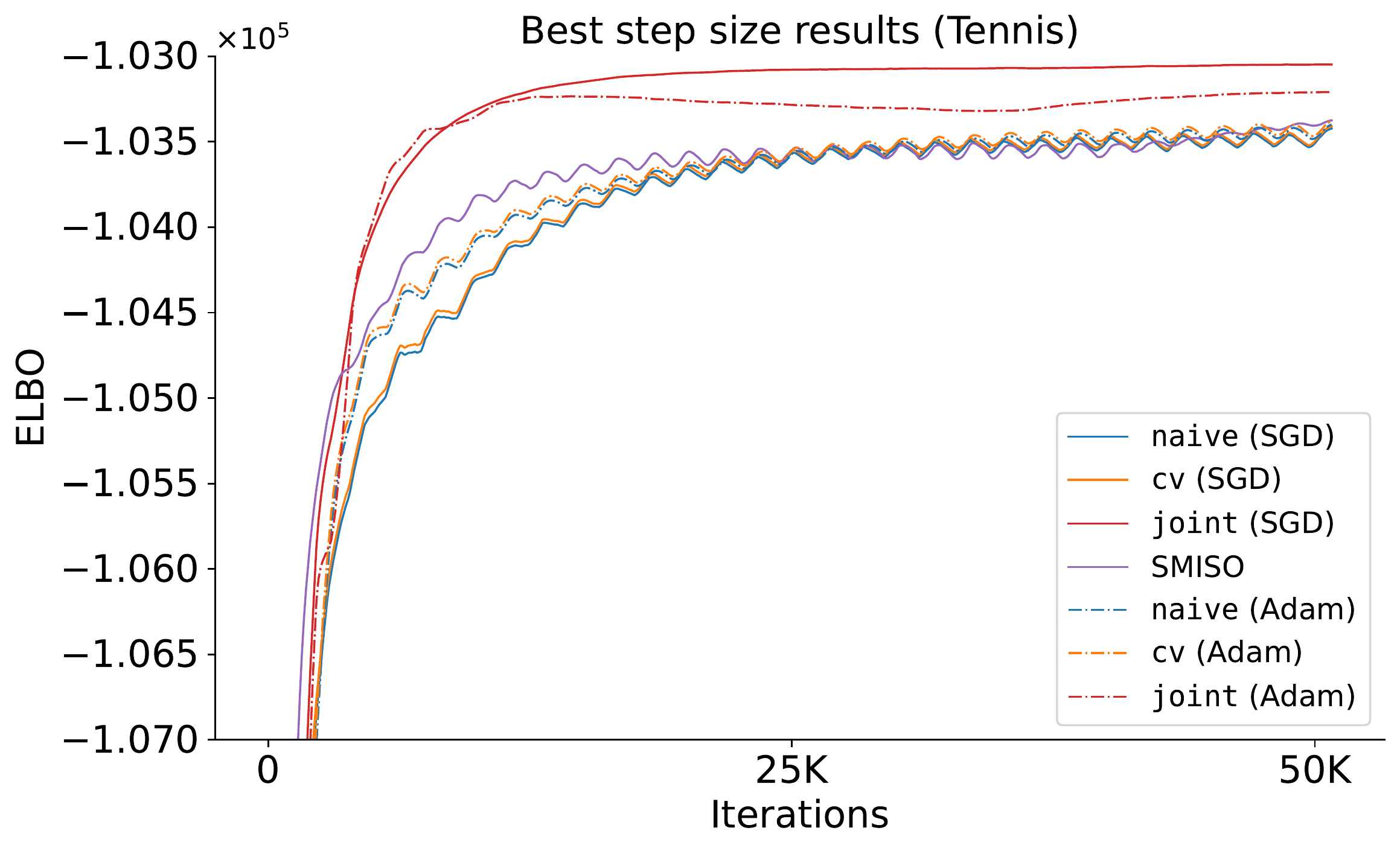}
  \label{fig:tennis_sgd_adam}   
\end{subfigure}
\begin{subfigure}{0.485\textwidth}
    \centering
       \includegraphics[width=\textwidth]{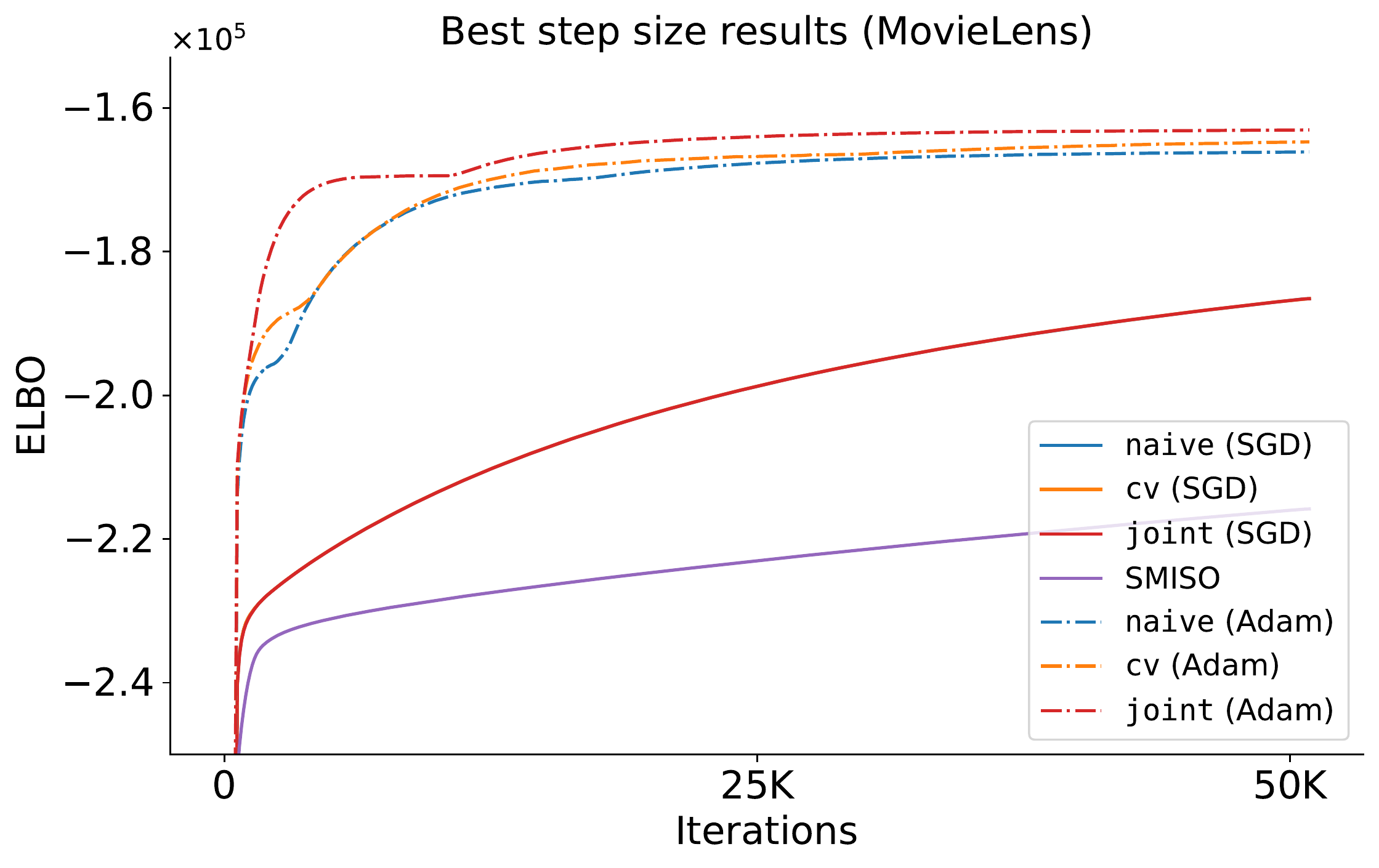}
  \label{fig:movie_sgd_adam}   
\end{subfigure}
    \caption{\textbf{Comparision of different estimators on MNIST, PPCA, and Tennis under SGD and Adam}. The proposed $\dual$ combined with Adam shows the best performance on all tasks except Tennis, in which $\dual$ with SGD demonstrates the best convergence. For other estimators, Adam leads to better and faster convergence than SGD.}\label{fig:sgd_vs_adam}
\end{figure}

\begin{figure}[ht]
\label{fig:full_results_sgd_big_row}
    \centering
    \begin{subfigure}{0.95\textwidth}
        \centering
        \includegraphics[width=\textwidth]{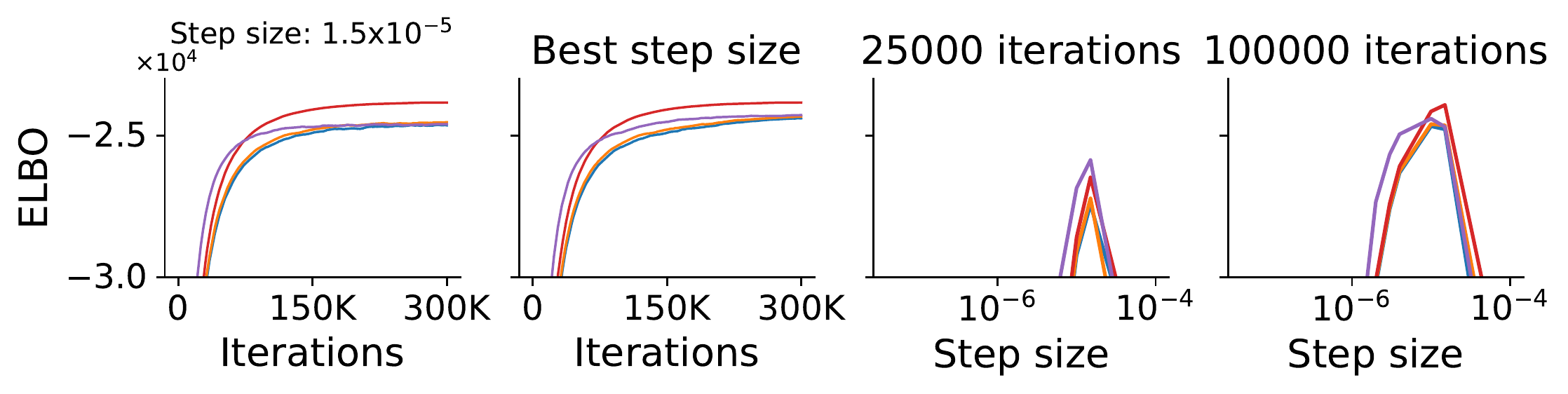}
        \caption{MNIST}
    \end{subfigure}
    \begin{subfigure}{0.95\textwidth}
        \centering
        \includegraphics[width=\textwidth]{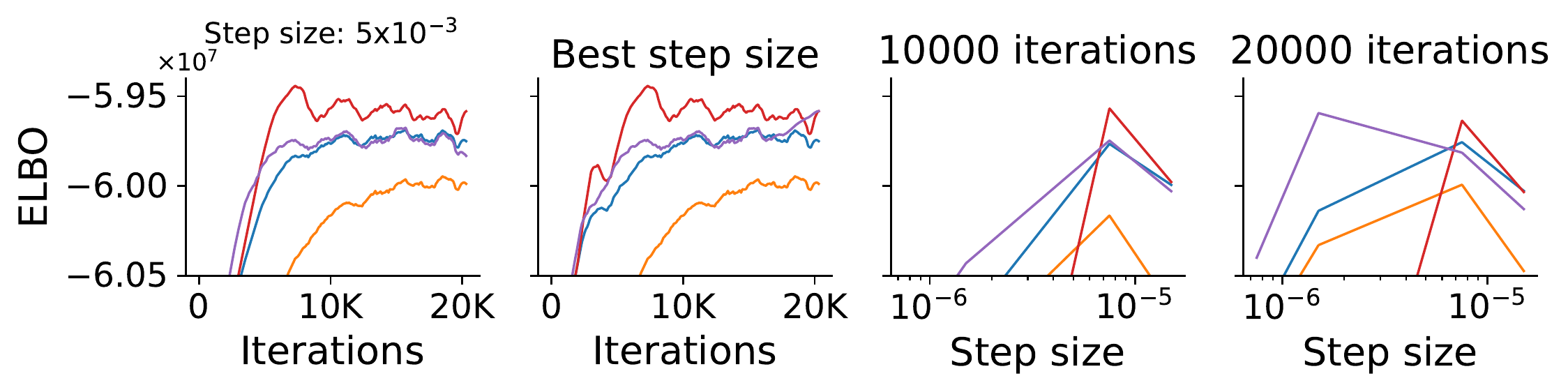}
        \caption{PPCA}
    \end{subfigure}
\begin{subfigure}{0.95\textwidth}
       \includegraphics[width=\textwidth]{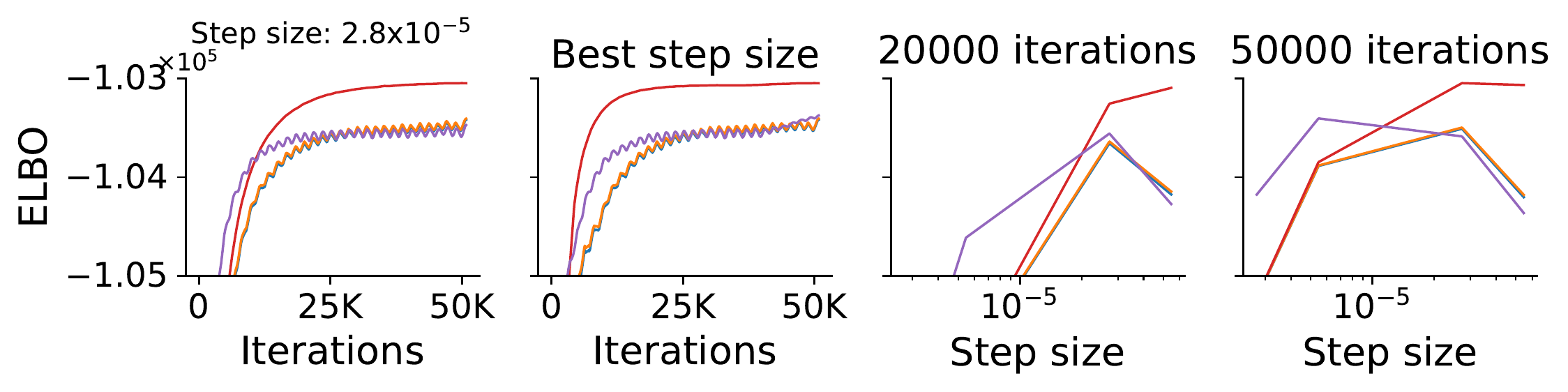}
       \caption{Tennis}
\end{subfigure}

\begin{subfigure}{0.95\textwidth}
       \includegraphics[width=\textwidth]{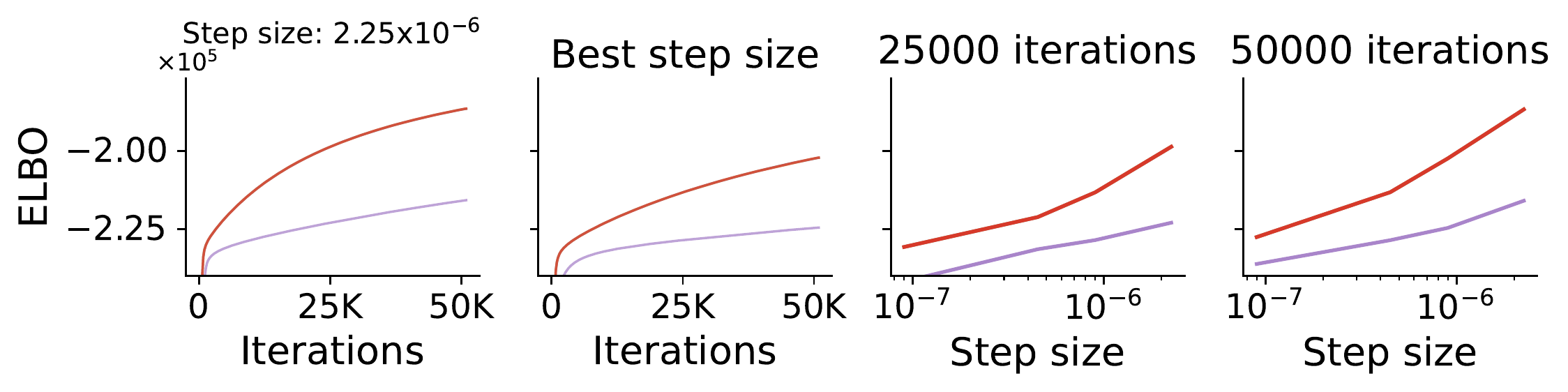}
       \caption{Tennis}
\end{subfigure}
    \centering
    \begin{subfigure}[b]{0.5\linewidth}
        \centering
        \includegraphics[width=\linewidth]{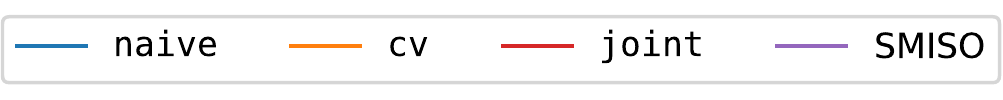}
        \caption*{} 
    \end{subfigure}
    \caption{\textbf{Optimization results on MNIST, PPCA, Tennis and MovieLens with SGD}. Using SGD does not affect the improvement of $\dual$ against $\naive$ and $\cv$. In addition, we notice that $\dual$ still performs better than SMISO under SGD, we suspect that this is because $\dual$ marginalizes $\epsilon$ out explicitly while SMISO approximates the expectation using exponential averaging. All methods show slow convergence with SGD on MovieLens under the largest step size, emphasizing the importance of using adaptive optimization methods.}
    \label{fig:sgd_results_big_row}
\end{figure}

In this section, we compare $\naive$, $\cv$, and $\dual$ with SMISO using SGD. The step sizes for SMISO are the same as the values shown in Sec.~\ref{sec:step_size_range}.
The step sizes for other models under SGD are converted through Eq.~\eqref{eq:smiso_stepsize_convertion} correspondingly. Additionally, we compare their performance with the optimization results acquired using Adam. The results are presented in Fig.~\ref{fig:sgd_vs_adam} and Fig.~\ref{fig:sgd_results_big_row}. Overall, with SGD, $\dual$ still shows superior performance compared with baseline estimators except for MovieLens, where all estimators fail to converge under the selected step sizes (and using larger step sizes could cause divergence in optimization). In addition, all estimators show performance worse than that of Adam when optimized with SGD except for $\dual$ on Tennis.

Note that, when experimenting with PPCA using $\dual$ and SGD, we perform updates with $\naive$ in the first three epochs to avoid diverging, as the $\dual$ shows a high gradient norm in the first few epochs when SAGA is still warming up. This modification is not required when using Adam, as Adam adaptively chooses the step size based on the gradient norm.

\section{Wall clock time v.s convergence}\label{sec:time}

\begin{figure}[ht]
    \centering
    \begin{subfigure}{0.44\textwidth}
        \centering
        \includegraphics[width=\textwidth]{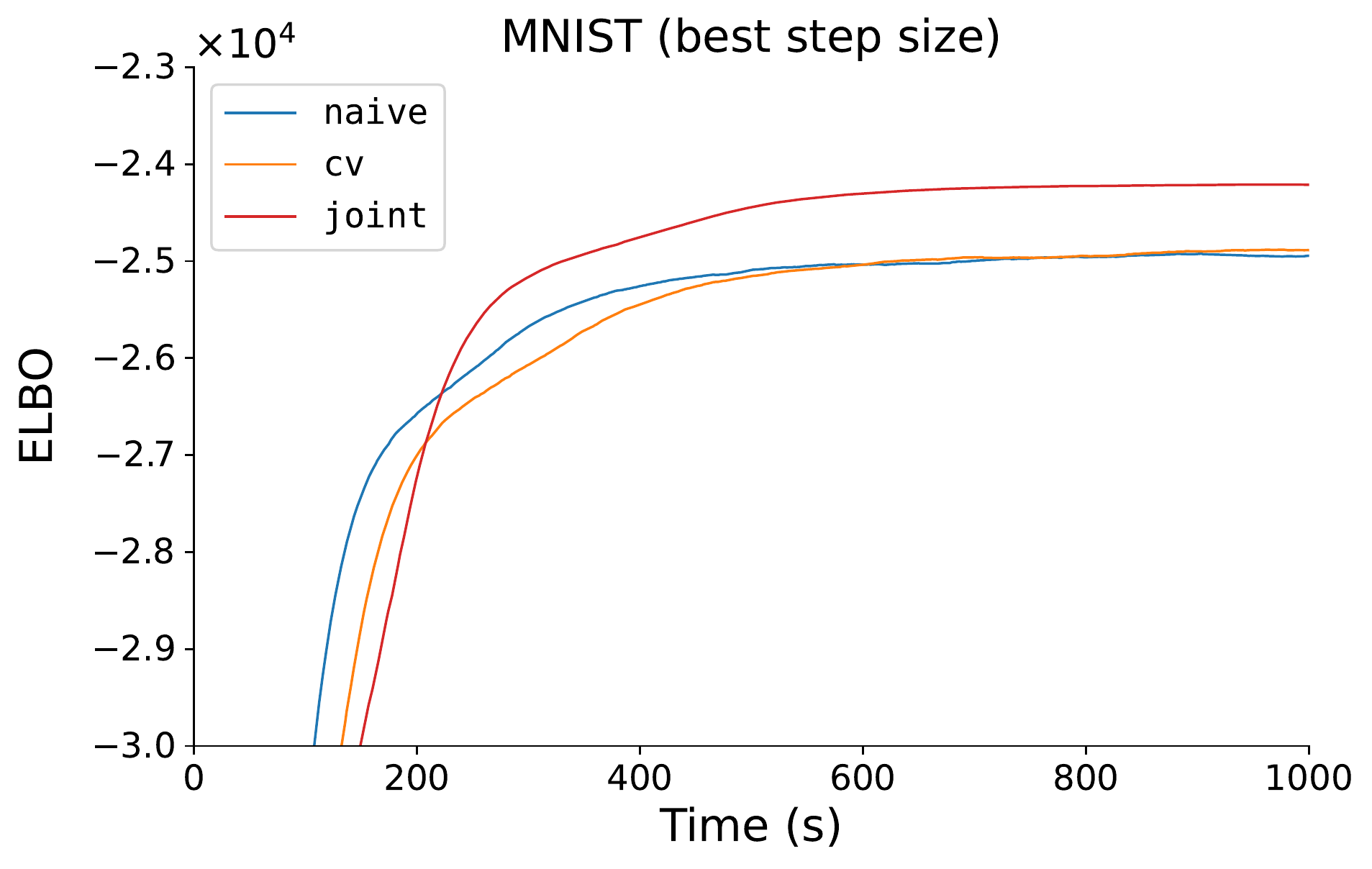}
    \end{subfigure}
    \begin{subfigure}{0.45\textwidth}
        \centering
        \includegraphics[width=\textwidth]{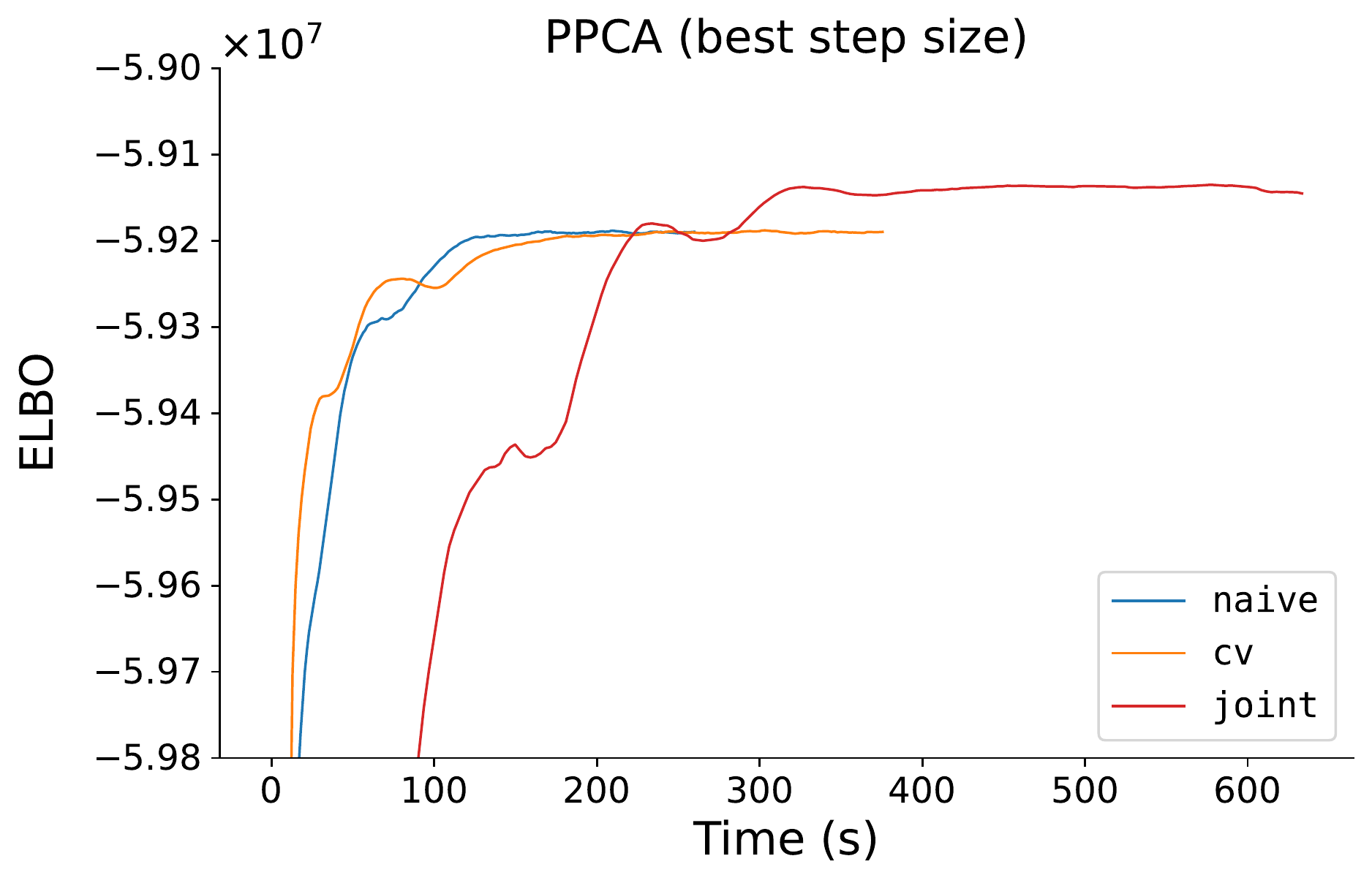}
    \end{subfigure}
\begin{subfigure}{0.45\textwidth}
    \centering
       \includegraphics[width=\textwidth]{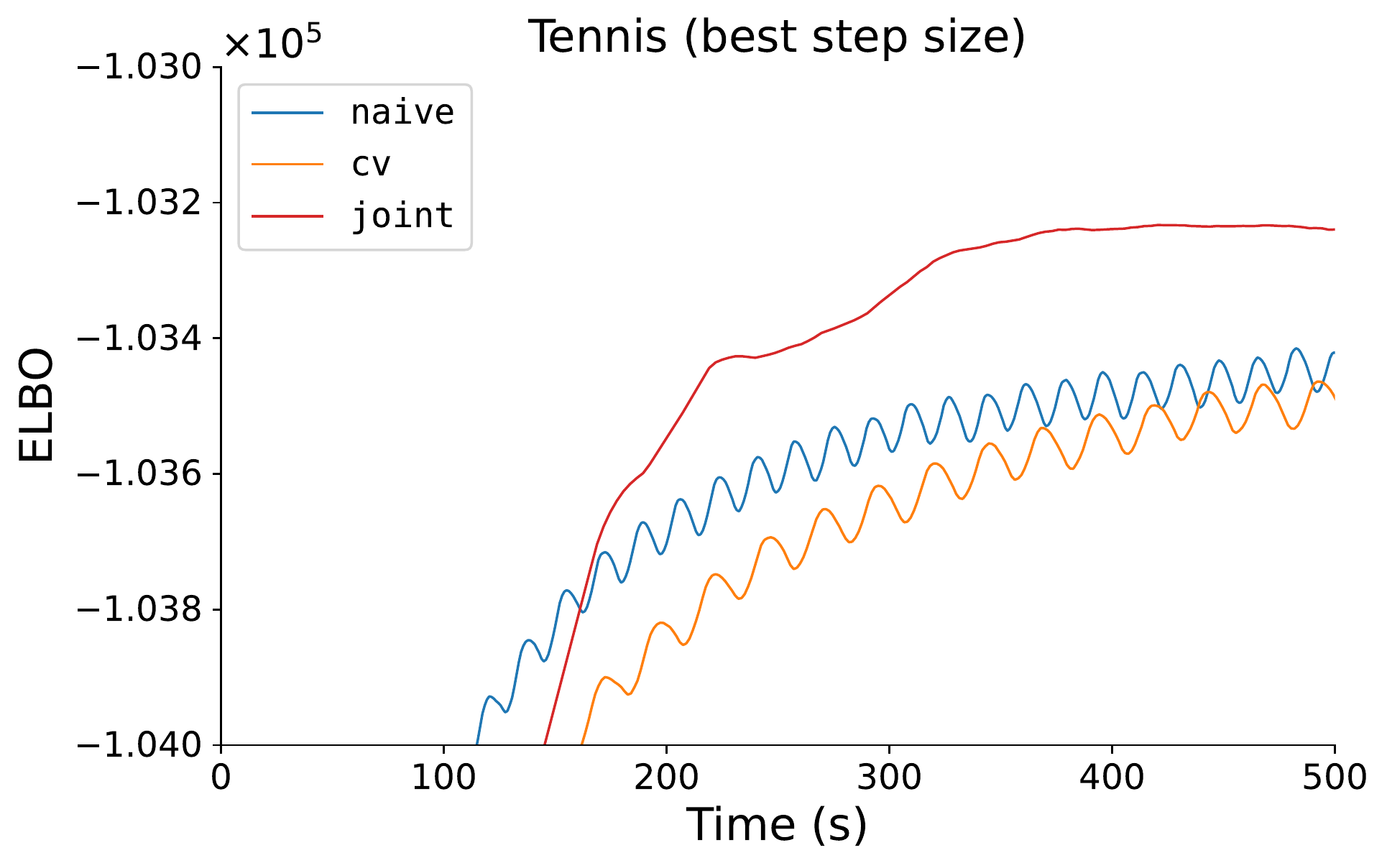}
\end{subfigure}
\begin{subfigure}{0.45\textwidth}
    \centering
       \includegraphics[width=\textwidth]{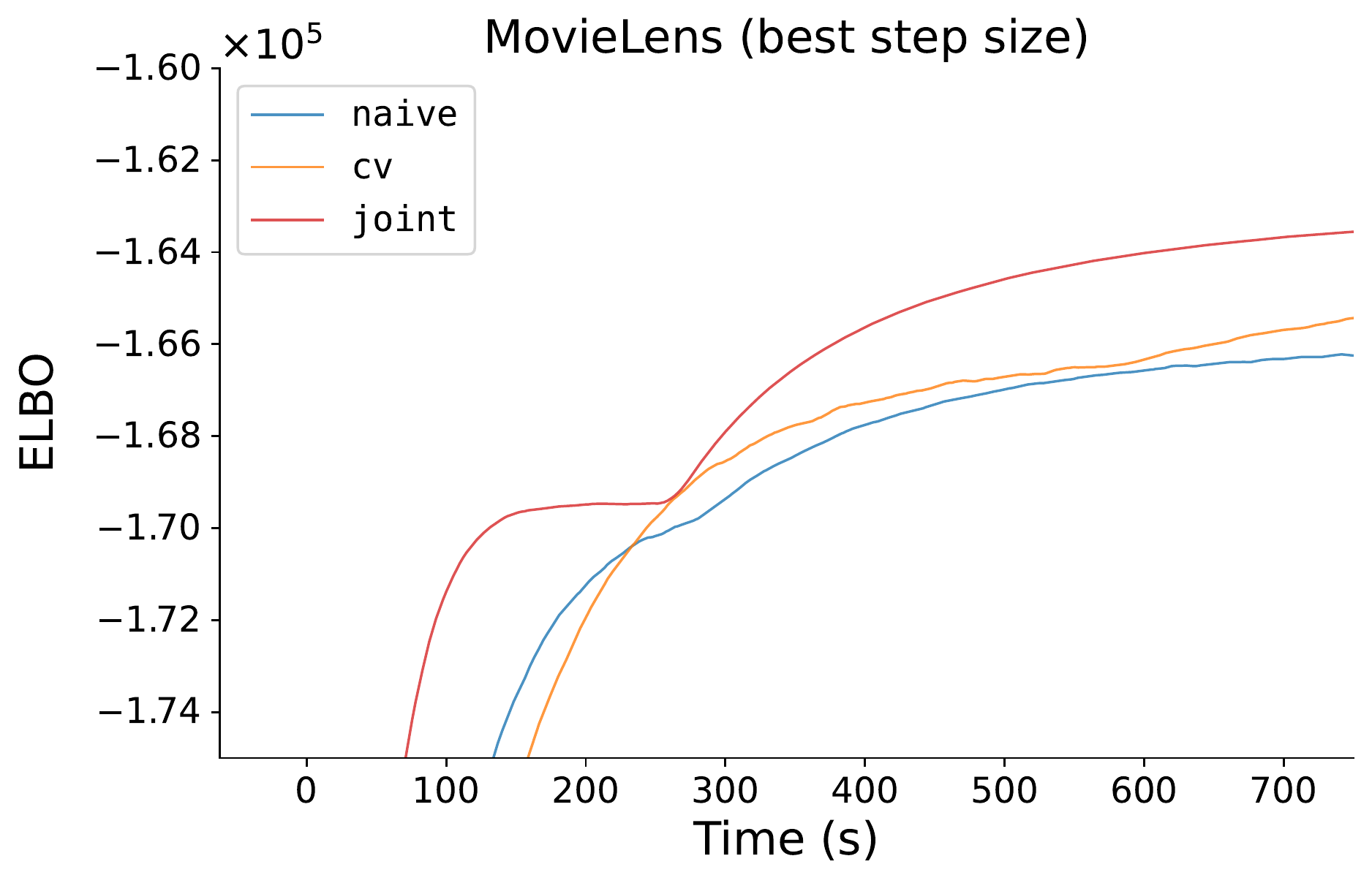}
\end{subfigure}
    \caption{\textbf{On large scale problems, the joint estimator leads to faster convergence in terms of wall-clock time.} For example, on MNIST, it takes $\dual$ around 300 seconds to reach an ELBO of $-2.5\times 10^{4}$ whereas the cv and the naive estimator would take around 600 seconds. On PPCA, while the $\dual$ estiamtor displays slower convergence in the beginning, as SAGA is still warming up, it is capable of reaching much better results at the end of the optimization.}
    \label{fig:time_elbo}
\end{figure}

In this section, we provide the wall clock time v.s. convergence results. The results are presented in Fig.~\ref{fig:time_elbo}. The results are identical to the results in the second column in Fig.~\ref{fig:big_results} with the x-axis for each estimator rescaled using the values from Table.~\ref{table:estimator_summary}.

\end{document}